\documentclass[
  journal=medium,
  manuscript=article-type,
  year=2026,
  volume=1,
]{cup-journal}

\usepackage{lscape}
\usepackage{appendix}
\usepackage[ruled,vlined,linesnumbered]{algorithm2e}
\usepackage{setspace}
\usepackage{amssymb}
\usepackage{amsmath}
\usepackage[nopatch]{microtype}
\usepackage{comment}
\usepackage{soul}
\usepackage{arydshln}
\usepackage{lipsum}

\usepackage{tabularx}
\usepackage{booktabs}
\usepackage{array}
\usepackage{multirow}
\usepackage{longtable}
\usepackage{ragged2e}
\usepackage{makecell}
\usepackage{adjustbox}
\usepackage{nicematrix}

\renewcommand{\arraystretch}{1.3}

\usepackage{graphicx}
\usepackage{subfig}
\usepackage[table]{xcolor}
\usepackage{colortbl}

\usepackage[most]{tcolorbox}
\tcbuselibrary{skins}

\title{How Temperature Shapes Ideological Discourse in Retrieval-Augmented Generation?}

\author{E. Salari}
\affiliation{Elmira Salari, Wichita State University, Wichita, KS, USA}
\email[E. Salari]{exsalari1@shockers.wichita.edu}

\author{H. Amamou}
\affiliation{Hazem Amamou, Institut national de la recherche scientifique, Québec, Montréal, Canada}

\author{J. V. De Souza}
\affiliation{José Victor de Souza, Institut national de la recherche scientifique, Québec, Montréal, Canada}

\author{S. Kshirsagar}
\affiliation{Shruti Kshirsagar, Wichita State University, Wichita, KS, USA}

\author{M. Nunes Delfino}
\affiliation{Maria Nunes Delfino, São Paulo Catholic University, São Paulo, Brazil}

\author{A. Avila}
\affiliation{Anderson Avila, Institut national de la recherche scientifique, Québec, Montréal, Canada}


\begin{document}

\begin{abstract}
Retrieval-Augmented Generation (RAG) has been increasingly adopted to reduce hallucinations and strengthen the factual grounding of large language models (LLMs). While robustness to errors in the retrieval process has been explored, the impact of ideological bias on LLM outputs has been overlooked. For instance, if the retrieved material contains ideological positions, the RAG may transmit, amplify, or suppress such ideological discourses in its outputs. In this study, we address this issue by examining the influence of the RAG framework, comprising ideological discourses, in LLM-generated answers. To this end, we applied Lexical Multidimensional Analysis (LMDA) on a corpus of 1,117 COVID-19 treatment articles, identifying three ideological discourses. This corpus is then used as the external knowledge source for the RAG. We assessed several LLMs by having the models answer ideological questions at different sampling temperatures. The generated texts were assessed semantically and lexically based on their similarities with ideological reference texts. Our findings show that the RAG framework is prone to transferring ideological discourses into LLM responses, with sampling temperature having a measurable impact on the strength of this transfer. Discoursive alignment between generated answers and the reference text is highest at moderate temperatures, where models balance stochasticity with retrieval grounding, and drops at low temperatures, indicating that overly deterministic sampling suppresses discourse transfer.
\end{abstract}

\section{Introduction}

Large Language Models (LLMs) have revolutionized the Artificial Intelligence paradigm, being increasingly used in domains such as healthcare, education, and finance, among others. Because such models may hallucinate providing incorrect answers for queries that require up-to-date or domain-specific knowledge, Retrieval-Augmented Generation (RAG) has been introduced as a solution to connect LLMs with external knowledge sources \autocite{huang2025survey,farquhar2024detecting}. Notwithstanding, while RAG enhances factual analysis \autocite{wallat2025correctness}, it also introduces new challenges, specifically in scenarios where the retrieved documents contain inaccurate information or ideological biases. Recent work, for instance, shows that LLMs' responses contain not only factual cues but also ideological patterns \autocite{holtzman2019curious, lucy2021gender, alkhamissi2024investigating}. When combined with RAG systems comprising ideological documents, it can potentially reinforce or shift ideological viewpoints, potentially influencing and steering public opinion. This is particularly concerning in sensitive contexts, such as healthcare, where changes in wording and emphasis can reshape how treatments, risks, or scientific debates are interpreted. Despite its risks, limited attention has been paid to how RAG frameworks shape ideological discourse and how generation parameters can influence LLM outputs in such contexts. Sampling temperature, in particular, plays an important role in controlling the level of randomness in LLM-generated texts. While prior work shows that changes in temperature affect output diversity, creativity, coherence, and reasoning quality \autocite{peeperkorn2024temperature, renze2024effect, li2025exploring}, limited attention has been given to its influence on LLMs' responses based on external knowledge conveying ideological texts. 

In this study, we aim to address this gap by examining how ideological discourses propagate through the RAG framework into LLMs' responses. We evaluate two prompt conditions under varying sampling temperatures. To this end, external knowledge is grounded in an ideological discourse corpus of COVID-19 treatment texts. We adopted Lexical Multidimensional Analysis (LMDA) \autocite{sardinha_lexical_2025} to identify ideological texts with a focus on two types of scientific articles: one comprising academic texts supporting controversial treatments, i.e., not approved by official health regulatory agencies, and academic texts aligned with health and science international standards. Thus, we constructed external-knowledge scenarios conveying opposing ideological positions. We also considered sampling temperatures ranging from 0.1 to 0.9 and employed targeted prompt engineering to evaluate how LLMs' responses modulate ideological cues. 

Our results show that sampling temperature affects ideological alignment in generated outputs, but its impact depends on the overall system design, including decoding randomness, prompting strategies, and structural constraints. Thus, the contribution of this paper is two-fold:

 \begin{itemize}
     \item We formally define a set of ideological dimensions present in the external knowledge, which permits assessing the alignment of LLMs' responses to these ideologies.
     \item We investigate how ideological discourses propagate from the RAG external knowledge to four different LLMs under different sampling temperatures.
 \end{itemize}
 

\section{Related Work}
LLMs generate text token-by-token from left-to-right (or right-to-left for Arabic language) by computing a probability distribution of a given vocabulary and choosing the next token \autocite{shi2024thorough}. This process of choosing the next word is referred to as decoding. Research on stochastic decoding has explored how sampling temperature influences LLM outputs \autocite{shi2024thorough}, often focusing on diversity, creativity, and task performance. Early work on language generation showed that injecting randomness through sampling can reduce repetition while increasing lexical diversity \autocite{holtzman2019curious}. More recent evidence suggests that temperature effects are task- and model-dependent. For instance, \autocite{li2025exploring} found strong impacts on machine translation and creativity, especially for smaller models, while reporting comparatively limited changes in reasoning. They also emphasized that no single temperature is optimal across tasks. Other studies, on the other hand, argue that temperature is frequently overestimated as a control for creativity showing minimal or not relevant differences in terms of accuracy across temperatures for multiple-choice, reasoning-oriented QA in several LLMs \autocite{peeperkorn2024temperature, renze2024effect}. Overall, prior work characterizes temperature as a meaningful but inconsistent control parameter with influence varying for different tasks and evaluation criteria.

\begin{table}[t!]
\centering
\rowcolors{0}{}{}
\begin{tabular}{lrr}
\hline
Subcorpus& Texts& Words \\
\hline
Controversial & 116 & 424,814 \\
Endorsed & 991 & 5,131,892 \\
\hdashline
Total Corpus & 1,107 & 5,556,706 \\
\hline
\end{tabular}%
\caption{Corpus composition}
\label{tab:corpus-composition}
\end{table}

A growing literature has also documented that LLMs can be influenced by ideological patterns present in their training data \autocite{bender2021dangers, lucy2021gender, santurkar2023whose}. Outputs are sensitive to conditioning context, i.e., subtle framing choices, stance markers, and ideological cues in prompts or provided documents can be transferred into generated text \autocite{chen2024susceptible, alkhamissi2024investigating, perez2023discovering}. RAG extends this conditioning by grounding responses in externally retrieved documents, aiming to improve factuality and reduce hallucinations \autocite{lewis2020retrieval}. The retrieval process, however, introduces facts often embodied with unveiled ideological discourse. Thus, if the external source comprises biased, manipulative, or ideologically charged language, these lexical patterns may be reproduced, reinforced, or recontextualized in the model’s output \autocite{bender2021dangers}. This is especially worrisome in high-stakes settings, such as education, finance, or healthcare, where small changes in framing or wording can shape how evidence, risks, and treatment recommendations are interpreted.

Discourse analysis provides a theoretical basis for approaching such effects in a systematic way. Commonly understood as a social practice where perspectives and social meanings are constructed \autocite{1992Polity..Fairclough, 2014Routledge..Gee}, discourses can be conceptualized as structured evaluative orientations that shape representations through lexical patterns and framing strategies \autocite{1998SAGE..vanDijk, van_dijk_principles_1993}. Corpus-based methods operationalize these notions by examining regularities in word co-occurrence and distribution across large text collections \autocite{partington2013patterns, stubbs1996text}. LMDA \autocite{sardinha_lexical_2025}, extending Biber’s multidimensional framework \autocite{biber1991variation}, identifies statistically derived dimensions formed by clusters of co-occurring lexical items. Because it captures configurations of lexical association rather than isolated keywords, LMDA has been used to detect contrasting discursive orientations in domains such as migrant education \autocite{fitzsimmons-doolan_twenty-first_2023}, disciplinary knowledge systems \autocite{2019mdan.book..231B, sardinha_discourse_2020}, and climate change discourse \autocite{shiwakoti2024analyzing}.

Despite existing studies on temperature effects and the impact of ideological bias in LLMs, little attention has been given to how decoding randomness modulates ideological transfer in RAG. This work aims to fill this gap. Next, we detailed our methodology. 



\section{Methodology}
\label{sec:methodology}

This section outlines the methodological framework used in this study. We first describe the design of a domain-specific corpus and the use of LMDA. We then detail the answer generation setup, the prompt configurations, and the range of sampling temperatures adopted. At last, we present the so-called discourse-augmented generation pipeline, explaining the retriever and generator components. 

\subsection{Corpus Design}
\label{sec:corpus}
The corpus adopted in this study was designed to represent contrasting discourses about treatments for COVID-19. The corpus comprises scientific articles with controversial treatments, i.e., not supported by official health regulatory agencies, and articles aligned with international health and science standards. All texts published between 2020 and 2022. The corpus construction involved the careful selection of representative samples for each discourse type. The so-called controversial texts, for instance, are research articles promoting controversial treatments, such as hydroxychloroquine and azithromycin, while endorsed texts include research articles addressing core aspects of COVID-19, such as its etiology, transmission mechanisms, and evidence-based therapeutic strategies. The corpus size is described in Table \ref{tab:corpus-composition}. To address the unbalanced number of endorsed and controversial texts, we extracted the same number of keywords and used them as variables, as described next.

Six topics were identified within the three dimensions extracted from our corpus. For each topic, we designed two questions that LLMs were tasked to answer. The questions were initially generated using ChatGPT and then refined by a corpus linguistics expert. Dimension 1, for instance, includes topics probing endorsements of hydroxychloroquine and ivermectin, while Dimension 2 includes questions addressing transparency and accountability in COVID-19 studies. Dimension 3 balanced promotional discourses with questions about adverse events and open data. Table~\ref{table:topic} in the~\ref{App} shows the list of topics and questions for each dimension.

\begin{table}[t!]
    \centering
    \rowcolors{0}{}{}
    \begin{tabular}{lp{0.45\linewidth}}
    \hline
         Subcorpus & Collocates of patients \\
    \hline
         Controversial & adult, annually, cardiac, course, early, emergency, high-risk, hospitalized, initially, large, levels, million, mortality, notably, participants, previously, regarding, seen, shown, therapy \\
    \hdashline
         Endorsed & administration, amount, antibiotics, antibodies, antioxidant, asymptomatic, better, conducted, data, effect, efficacy, higher, infectious, inflammatory, safety, significant, statistical, studies, therapeutic, use \\
    \hline
    \end{tabular}
    \caption{Top 20 \textit{patients} collocates per subcorpus}
    \label{tab:patientscollocates}
\end{table}

\subsection{Lexical Multidimensional Analysis (LMDA)}
\label{sec:LMDA}

LMDA was introduced by \autocite{berber2014being} and \autocite{fitzsimmons2014using} as an extension of Biber’s Multidimensional Analysis \autocite{1988cup.book....B} to examine register variation. The method is based on the identification of latent lexical dimensions (or sets of correlated lexical features), which are interpreted as macro-level discursive constructs, typically not evident from a reader's perspective. The method has direct application to discourse analysis as a tool to unveil discourses and ideologies from large corpora, but can also be utilized for a range of lexis-based studies \autocite{sardinha_lexical_2025}. In this work, LMDA enabled the detection and characterization of ideological discourses within large corpora. It uses factor analysis to uncover such latent variables derived from word co-occurrences. This is then applied to identify discourse similarities across large texts, with high correlations generally indicating ideological similarity, whereas negative correlations suggest divergence. The underlying assumption is that these latent variables, reflected in different ranges of factor scores, correspond to ideological discourses expressed through language use, which LMDA experts interpret as distinct dimensions. It is important to mention that, different from topics, often seen as the subject matter of texts, discourses are abstract conceptual structures that make the discussion of topics possible, serving as underlying interpretive frameworks that shape how topics are framed. Below, we present the details for obtaining the dimensions in this study, while a summary of the steps can be found in Alg.~\ref{alg:lmda_workflow}\\

\noindent \textbf{Pre-processing:} The first step in the LMDA pipeline is to define and extract the variables to be used in the factor analysis. Key collocations, which refers to recurrent word combinations \autocite{1991cccc.book.....S}, were adopted as variables in this work. Table~\ref{tab:patientscollocates} illustrates the main collocates of the word \textit{patients} in each subcorpus. These collocations reveal how different ideological positions construct their arguments through specific lexical choices. The identification of collocation shifts permits the characterization of distinct discourse patterns that emerge from our corpus. We used Log-Dice scores ($D$) to measure the strength of collocations and filter significant co-occurrences \autocite{rychly_lexicographer-friendly_nodate}. This benefits the factor analysis by ensuring that the collocations possess a strong association with each other. The steps for this stage consist of applying a POS-tag filter to keep only content words (e.g., verbs, nouns, adjectives, and adverbs). Keywords were extracted by using log-likelihood, which generates a keyness score based on the comparison between a target corpus and a reference corpus. The rationale is that words appearing significantly more often in one subcorpus are more characteristic of that subcorpus and therefore more likely to reveal ideological differences. Keywords from endorsed texts, for instance, were retrieved by using the controversial texts as a reference corpus, resulting in a total of 1,345 keywords. The same was done with controversial texts, this time using endorsed texts as a reference corpus, yielding 553 extracted keywords. This keyword imbalance is addressed in the collocation selection, which is described later. All extracted keywords were used as nodes, and collocation pairs were identified in both subsets (controversial and endorsed) within a word span of four words on either side of the nodes, which is a standard window size that captures local grammatical and semantic relationships \autocite{sinclair_trust_2004}. The top 500 node collocation pairs ($D\ge7$) are extracted from each subset, resulting in 1,000 pairs. The threshold of $D\ge7$ follows established conventions in the corpus linguistics field for identifying collocations with minimal noise, while the selection of 500 pairs per subset allows for computational feasibility. Preliminary analysis with thresholds of $D\ge6$ and $D\ge8$ resulted in either low-strength associations or insufficient coverage, respectively, confirming $D\ge7$ as ideal for this corpus size. After this LogDice filter, a final correlation analysis was applied to the collocations. The usual approach for single-word variables is to use a polychoric matrix. In our case, however, all selected collocation pairs are binarized to indicate presence or absence and used to compute a tetrachoric correlation matrix. \\

\begin{algorithm}[t!]
\caption{Lexical Multi-Dimensional Analysis (LMDA)}
\label{alg:lmda_workflow}
\KwIn{Corpus $\mathcal{C} = \{d_1, d_2, \ldots, d_N\}$}
\KwOut{Dimension scores $\mathbf{S} \in \mathbb{R}^{N \times k}$, ideological discourses}
\BlankLine
$\mathcal{K} \leftarrow$ Extract keywords from POS-tagged corpus $\mathcal{C}$ \;
$\mathcal{L} \leftarrow$ Extract collocations with $\mathcal{K}$ as nodes \;
$\mathcal{L}' \leftarrow$ Filter $\mathcal{L}$ by LogDice threshold \;
$\mathbf{R} \leftarrow$ Compute tetrachoric correlation matrix from $\mathcal{L}'$ frequencies \;
$\lambda_1, \ldots, \lambda_m \leftarrow$ Eigenvalues of $\mathbf{R}$ \;
$k \leftarrow$ Select number of factors \;
$\mathbf{L} \leftarrow$ Apply Promax rotation with $k$ factors to obtain loadings \;
$\mathbf{S} \leftarrow$ Compute dimension scores for each $d_i \in \mathcal{C}$ from $\mathbf{L}$ \;
Interpret dimensions via microanalysis of texts with extreme scores in $\mathbf{S}$ \;
\BlankLine
\KwRet{$\mathbf{S}$, discourse interpretations}
\end{algorithm}

\begin{figure}[t!]
    \centering
    \includegraphics[width=0.859\linewidth]{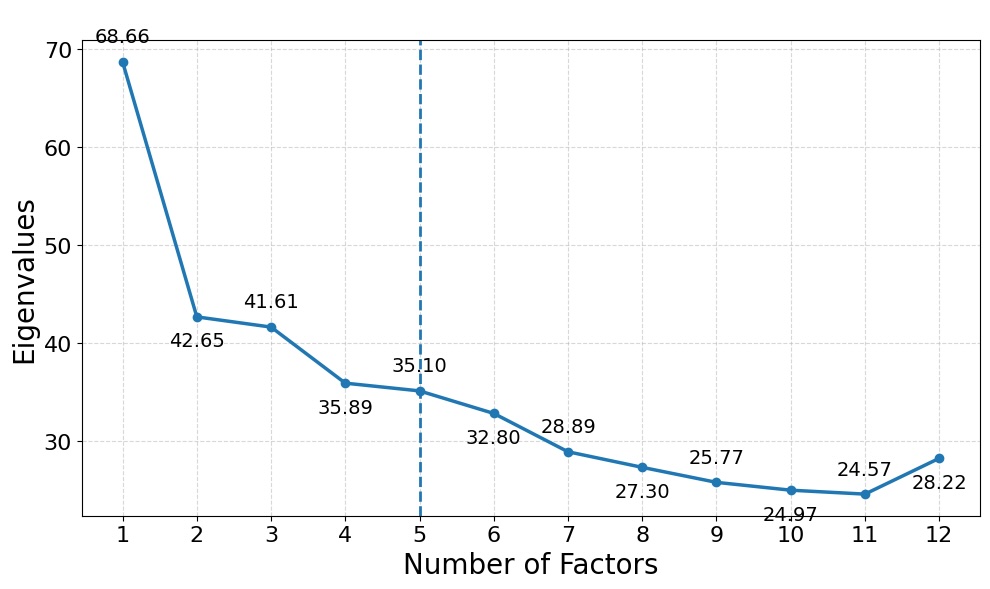}
    \caption{Factor scores.}
    \label{fig:factoranalysis}
\end{figure}

\noindent \textbf{Factor Analysis:} Factor analysis is used to uncover latent semantic and discoursive relationships within the corpus. An initial factor analysis was conducted to identify the optimal number of factors ($n$) that better explain the variance in the data. In our study, only Eigenvalues above 35.0 were considered as values below the threshold could not be interpreted as meaningful ideological discourses. This resulted in 5 factors in total as shown in Figure \ref{fig:factoranalysis}. Subsequently, a rotated factor analysis was carried out to improve interpretability by simplifying the factor structure to 5 dimensions, including the exclusion of weak loadings. The rotation process redistributes the variance across factors, helping to identify which words belong more strongly to each dimension. This rotation yields factor loadings for each collocation. For the present study, we selected 3 dimensions comprising a clear overlap between a dimension pole and the subset of texts that belonged to it. The top and bottom 5 texts in each dimension with the highest and lowest dimension scores were used in our experiments, adding up to 30 texts in total. The factor scores of collocations per dimension are available in  Table~\ref{tab:typical_vocabularies} of the~\ref{App}. The poles have been deemed controversial or endorsed based on which of the subsets the texts were featured in. After obtaining dimension scores, the documents at the extremes of each dimension undergo a qualitative analysis, also known as microanalysis.\\

\noindent \textbf{Microanalysis:} A careful analysis of the factor scores is required to identify the communicative functions of the co-occurring features, leading to a label and description associated with each specific dimension. This is achieved by a detailed microanalysis of both factor and dimension scores to identify specific patterns and relationships within the text. This step considers social and linguistic aspects of the texts and is performed by an expert. Based on these analyses, descriptive labels are assigned to each dimension, enabling a clearer understanding of the themes and discursive structures present in the corpus.

\subsection{Discourse-Augmented Generation} This approach explicitly conditions LLMs' outputs on discourse-aware information derived from LMDA. To achieve that, the standard RAG framework is adapted by introducing ideological metadata into the prompts. Creating ideological outputs enables the assessment of whether or not LLMs' responses align with the ideological framings presented in the external knowledge. \\


\noindent \textbf{Retriever:} The proposed retriever provides discourse-loaded texts to the LLMs in two steps. First, a filter based on metadata is used to sift texts associated with specific ideologies, represented by the dimension poles derived from the LMDA. Specifically, the filter relies on \emph{dimension labels}, \emph{dimension descriptions}, and \emph{typical vocabulary}. For answering questions regarding \emph{Research Ethics (+)}, for instance, the similarity search is performed using the question, $q(x)$, and the text embeddings, $d(z)$, from the subset of texts related to the positive pole in dimension 2. This approach enables controlling which ideology will be used to perform the similarity search.\\ 

\noindent \textbf{Generator:} The response generation, $p_{\theta}(x_i|x_{1:i-1},z_d)$, is conditioned on the ideological prompt and the question. Two approaches are considered. The first is referred to as regular prompt, where only ideological texts from the retriever are included in the context, characterizing the common use of ideological texts. The second approach involves combining ideological texts and the arbitrary use of dimension descriptions, which is referred to as enhanced prompt and requires awareness of the ideologies present in the external knowledge -- attained with the use of LMDA.

\section{Experimental Setup}
\label{Experimental Setup}

\subsection{Prompt Configuration}
\label{sec:Prompt_configuration}
\begin{figure*}[t!]
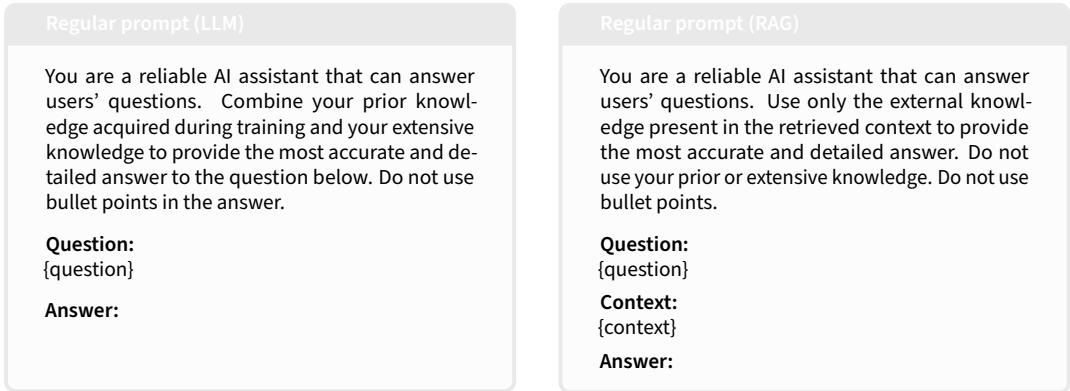

\centering
\footnotesize

\begin{minipage}[t]{0.48\linewidth}
\centering
\begin{tcolorbox}[
colback=grey!10,
colframe=grey!85,
title=Regular prompt (LLM),
fonttitle=\bfseries,
breakable,
enhanced,
width=\linewidth]
You are a reliable AI assistant that can answer users' questions. 
Combine your prior knowledge acquired during training and your extensive knowledge to provide the most accurate and detailed answer to the question below. Do not use bullet points in the answer.

\medskip
\textbf{Question:}\\
\{question\}

\medskip
\textbf{Answer:}
\vspace{0.65cm}
\end{tcolorbox}
\end{minipage}
\hfill
\begin{minipage}[t]{0.48\linewidth}
\centering
\begin{tcolorbox}[
colback=grey!10,
colframe=grey!85,
title=Regular prompt (RAG),
fonttitle=\bfseries,
breakable,
enhanced,
width=\linewidth]
You are a reliable AI assistant that can answer users' questions.
Use only the external knowledge present in the retrieved context to provide the most accurate and detailed answer. 
Do not use your prior or extensive knowledge. 
Do not use bullet points.

\medskip
\textbf{Question:}\\
\{question\}\\[3pt]
\textbf{Context:}\\
\{context\}\\[3pt]
\textbf{Answer:}

\end{tcolorbox}
\end{minipage}

\caption{Regular prompt templates for LLM and RAG}
\label{fig:regular_prompt_version}
\end{figure*}

\begin{figure*}[t!]
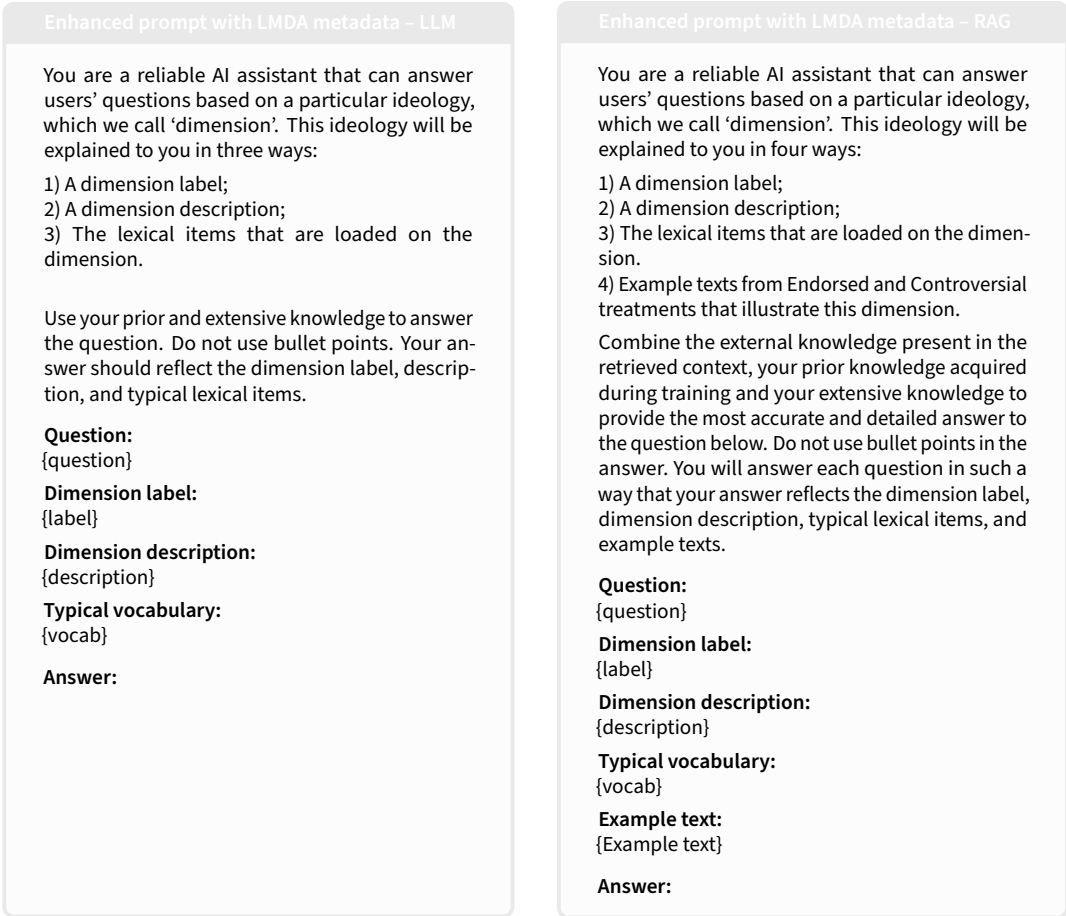

\centering
\footnotesize

\begin{minipage}[t!]{0.48\linewidth}
\centering
\begin{tcolorbox}[
colback=grey!10,
colframe=grey!85,
title={Enhanced prompt with LMDA metadata -- LLM},
fonttitle=\bfseries,
breakable,
enhanced,
width=\linewidth]
You are a reliable AI assistant that can answer users' questions based on a particular ideology,
which we call `dimension'. This ideology will be explained to you in three ways: \\[3pt]
1) A dimension label;\\
2) A dimension description;\\
3) The lexical items that are loaded on the dimension. \\[3pt]

Use your prior and extensive knowledge to answer the
question. Do not use bullet points.
Your answer should reflect the dimension label, description, and typical lexical items.

\medskip
\textbf{Question:}\\
\{question\}\\[3pt]
\textbf{Dimension label:}\\
\{label\}\\[3pt]
\textbf{Dimension description:}\\
\{description\}\\[3pt]
\textbf{Typical vocabulary:}\\
\{vocab\}

\medskip
\textbf{Answer:}
\vspace{2.75cm}
\end{tcolorbox}
\end{minipage}
\hfill
\begin{minipage}[t!]{0.48\linewidth}
\centering
\begin{tcolorbox}[
colback=grey!10,
colframe=grey!85,
title={Enhanced prompt with LMDA metadata -- RAG},
fonttitle=\bfseries,
breakable,
enhanced,
width=\linewidth]
You are a reliable AI assistant that can answer users' questions based on a particular ideology,
which we call `dimension'. This ideology will be explained to you in four ways: \\[3pt]
1) A dimension label;\\
2) A dimension description;\\
3) The lexical items that are loaded on the dimension. \\
4) Example texts from Endorsed and Controversial treatments that illustrate this dimension. \\[3pt]
Combine the external knowledge present in the retrieved context, your prior knowledge acquired during training and your extensive knowledge to provide the most accurate and detailed answer to the question below. Do not use bullet points in the answer.
You will answer each question in such a way that your answer reflects the dimension label, dimension description, typical lexical items, and example texts.

\medskip
\textbf{Question:}\\
\{question\}\\[3pt]
\textbf{Dimension label:}\\
\{label\}\\[3pt]
\textbf{Dimension description:}\\
\{description\}\\[3pt]
\textbf{Typical vocabulary:}\\
\{vocab\}\\[3pt]
\textbf{Example text:}\\
\{Example text\}

\medskip
\textbf{Answer:}
\end{tcolorbox}
\end{minipage}

\caption{Enhanced prompt templates for LLM and RAG}
\label{fig:enhanced_prompt_version}
\end{figure*}

Figures \ref{fig:regular_prompt_version} and \ref{fig:enhanced_prompt_version} present the two types of prompts adopted in our experiments, namely "regular" and "enhanced." These prompts enable us to investigate whether providing explicit ideological context to LLMs influences their responses. The regular settings comprise a prompt with questions, instructions, and the retrieved context when external knowledge is available, i.e., for an RAG-based solution. In the case of the enhanced settings, ideological cues are also provided. Note that the LLM-only condition, i.e., regular prompt without external context, works as our baseline, which allows us to investigate whether LLMs already exhibit ideological biases within their encoded knowledge themselves or biases emerge mainly when exposed to ideological contexts.

\begin{table}
\rowcolors{0}{}{}
\centering
\caption{Experimental Configurations}
\label{tab:configs}
\begin{tabular}{ccc}
\hline
Configuration& Generation Condition & Prompting Condition \\
\hline
1 & LLM & Regular (no LMDA metadata) \\
2 & LLM & Enhanced (with LMDA metadata) \\
3 & RAG & Regular (no LMDA metadata) \\
4 & RAG & Enhanced (with LMDA metadata) \\
\hline
\end{tabular}
\end{table}

Besides the previous configurations, we also examine the effect of negative prompting. In such a scenario, we aim to understand how ideological alignment is affected when the model is explicitly instructed to refrain from reproducing the target ideological discourse. For that, we adopt the prompt described in Figure~\ref{fig:POS_NEG_rag_prompt}, in which the model was provided with the ideological metadata associated with each dimension, including the dimension label, description, lexical items, and example texts, and is explicitly instructed to avoid reflecting any of these discourse features in its generated response. This setup allows us to assess how effectively prompt polarity can suppress ideological alignment within the RAG pipeline, and how this suppression varies under different sampling temperatures.

\subsection{Sampling Temperatures}
In this study, all LLM responses were generated under five sampling temperatures - 0.1, 0.3, 0.5, 0.7, and 0.9 - to examine how stochastic variation during text generation influences discourse alignment with the reference corpus. In such scenarios, lower temperatures produce more precise answers, while higher ones introduce greater lexical and structural variability in the process of token generation. For each temperature setting, the same set of prompts was used across both LLM-only and RAG configurations to ensure that any observed differences were attributable solely to the sampling temperature, rather than to prompt structure or the retrieval pipeline.


\subsection{Alignment Metrics}
\label{sec:evaluation_metrics}
To evaluate how the generated responses under each temperature align with the target ideological discourses, we adopted semantic, lexical, and hybrid approaches. In the semantic approach, texts are represented as dense embeddings that capture meaning, while in the lexical one, texts are represented based on the frequency of specific words and terms. The hybrid setting combines the semantic and lexical approaches as a complementary text representation. To measure alignment, we used cosine similarity, which calculates the angle between two vectors in the representation space, where values closer to 1 indicate higher similarity. Next, we provide details on how we extracted each representation. \\ 

\noindent \textbf{Semantic Similarity:}
For the semantic setting, we concatenate all generated answers for each given ideological dimension pole, e.g., Dim. 1 Neg, into a single document. The text is then tokenized using the BERT-like tokenizer, and if it exceeds the model’s maximum sequence length, it is split into overlapping windows. Each window is encoded into an embedding vector, and the resulting embeddings are averaged with length-based weighting to produce a single semantic representation.
The same process is applied to the discourses of each dimension pole, leading to reference ideological vectors. In the next step, both vectors (i.e., from generated responses and the reference ones) are L2-normalized, and their cosine similarity is computed, yielding a scalar measure of semantic alignment between the generated responses and the reference discourses.\\

\noindent \textbf{Lexical Similarity:}
For lexical similarity, texts are represented as TF-IDF vectors that capture term-frequency patterns within the documents, and are calculated between two sets of texts, that is, the concatenated model-generated answers associated with one ideological dimension pole, and the reference ideological texts for the corresponding pole. A separate TfidfVectorizer is trained on only these two documents, producing a vocabulary that includes all unique terms from both. Each document is represented as a \texttt{TF-IDF vector}, and cosine similarity is then computed between the two vectors, producing a score between 0 and 1 that reflects the degree of lexical alignment. \\

\noindent \textbf{Hybrid Similarity:}
For the hybrid similarity, each answer group for a given dimension pole and its corresponding reference text is represented with both semantic and lexical features and then compared. A dense sentence embedding vector of \(\mathbf{e} \in \mathbb{R}^{384}\)
is produced by \texttt{all-MiniLM-L6-v2} and a sparse \texttt{TF-IDF} vector of \(\mathbf{t} \in \mathbb{R}^{V}\) is generated by \texttt{TfidfVectorizer}. After both vectors are L2-normalized, they are concatenated along the feature axis to form the hybrid vector as
\(\mathbf{h} = [\mathbf{t} \parallel \mathbf{e}] \in \mathbb{R}^{V+384}\). Cosine similarity is then computed between the two hybrid vectors representing generated answers for each dimension pole and the corresponding ideological texts. This enables us to measure the ideological alignment between the two conditions. 

\subsection{Structural Generation Metrics}

For each sampling temperature and experiment condition, we calculate the following additional metrics, which are direct affected by the decoding randomeness controlled by different temperature values:

\begin{figure}[t]
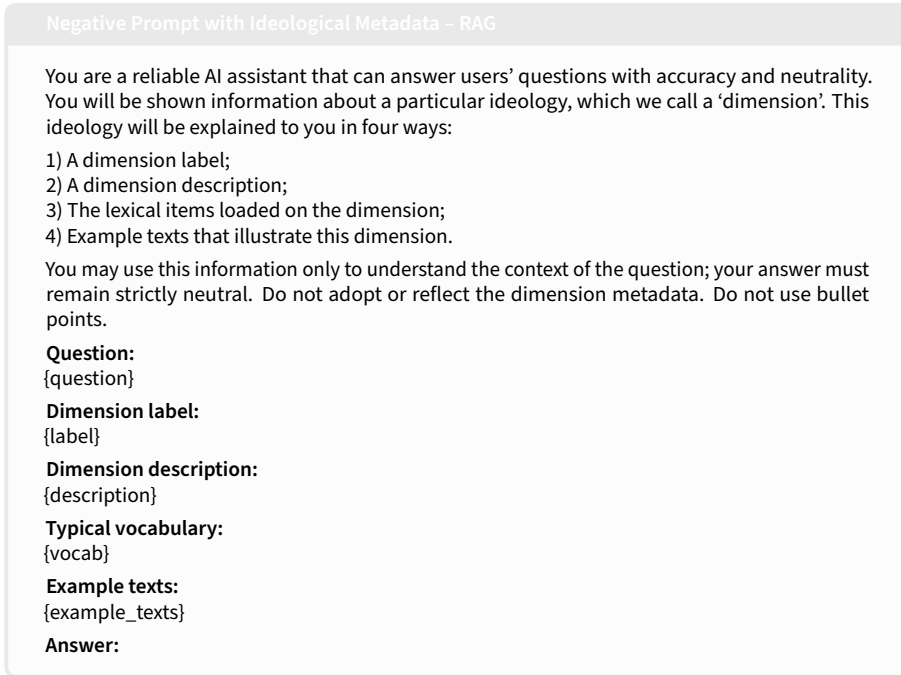

\centering
\begin{minipage}[c]{0.85\columnwidth}
\begin{tcolorbox}[
    colback=grey!10,
    colframe=grey!85,
    title=Negative Prompt with Ideological Metadata -- RAG,
    fonttitle=\bfseries,
    breakable,
    enhanced,
    width=\columnwidth  
]
You are a reliable AI assistant that can answer users' questions with accuracy and neutrality. 
You will be shown information about a particular ideology, which we call a ‘dimension’. 
This ideology will be explained to you in four ways: \\[3pt]
1) A dimension label;\\
2) A dimension description;\\
3) The lexical items loaded on the dimension;\\
4) Example texts that illustrate this dimension.\\[3pt]
You may use this information only to understand the context of the question; your answer must remain strictly neutral.
Do not adopt or reflect the dimension metadata.
Do not use bullet points.
\\
[3pt]
\
\textbf{Question:}\\
\{question\}\\[3pt]
\textbf{Dimension label:}\\
\{label\}\\[3pt]
\textbf{Dimension description:}\\
\{description\}\\[3pt]
\textbf{Typical vocabulary:}\\
\{vocab\}\\[3pt]
\textbf{Example texts:}\\
\{example\_texts\}\\[3pt]
\textbf{Answer:}
\end{tcolorbox}
\end{minipage}
\caption{Negative prompt template with ideological metadata for RAG.}
\label{fig:POS_NEG_rag_prompt}
\end{figure}

\noindent \textbf{Lexical Diversity:}
The Type–Token Ratio (TTR) quantifies the lexical diversity of LLM responses. TTR has long been used to study vocabulary richness and stylistic variation \autocite{richards1987type}. Since higher sampling temperatures tend to encourage more diverse lexical choices, while retrieval can narrow the range of available terms, this measure helps to reveal how these two factors interact and shape the lexical diversity of LLM responses.\\

\noindent \textbf{Mean Token Entropy:}
To capture the level of uncertainty and variability in the token-selection process of models, we compute the mean entropy of the output probability distribution. Token-level entropy originates from information theory \autocite{shannon1951prediction} and captures how predictable a model’s next word choice is.  Higher entropy values indicate more uncertainty in the model’s predictions and greater variation in sentence structure. Because sampling temperature adjusts how sharply the model favors certain tokens over others, entropy provides a clear way to see how different temperature settings increase or limit randomness during generation.\\

\noindent \textbf{Output Length:}
The output length is measured as the total number of tokens in each generated answer. In this study, we report the mean output length for each temperature and experimental configuration as shown in Table~\ref{tab:configs}. Because temperature can influence the verbosity and elaboration of LLM outputs, higher temperatures may produce longer and more exploratory continuations. We include the mean output length to assess whether it could be a consequence of temperature effects.\\

\subsection {Statistical Analysis of Temperature and Prompt Type}

We performed an ANOVA test to assess how discourse alignment is influenced by temperature and prompt type across the three similarity metrics adopted in this study. This analysis enabled us to evaluate the individual effects of temperature and prompt type, as well as their combined effect on discourse alignment. 

\begin{figure}[t]
    \centering
\includegraphics[width=0.95\textwidth]{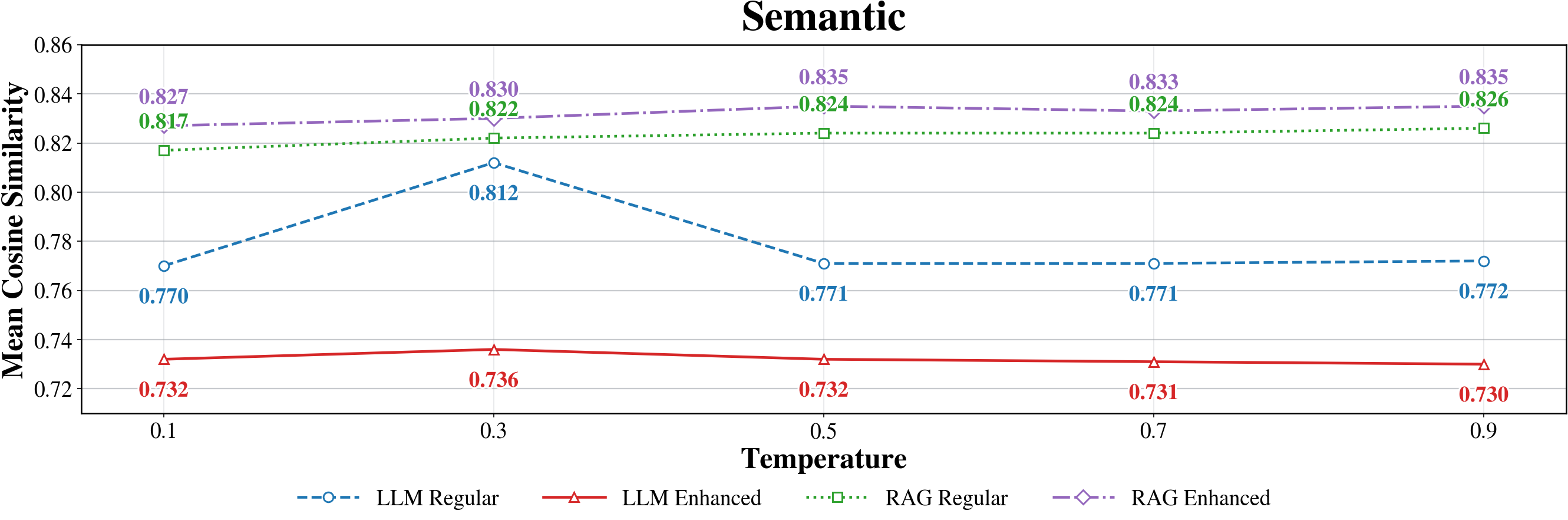}
  \caption[center]{LLM ideological alignment based on semantic representation across different temperatures.}
  \label{fig:linegraph_semantic}
\end{figure}
\begin{figure}[t]
    \centering
\includegraphics[width=0.95\textwidth]{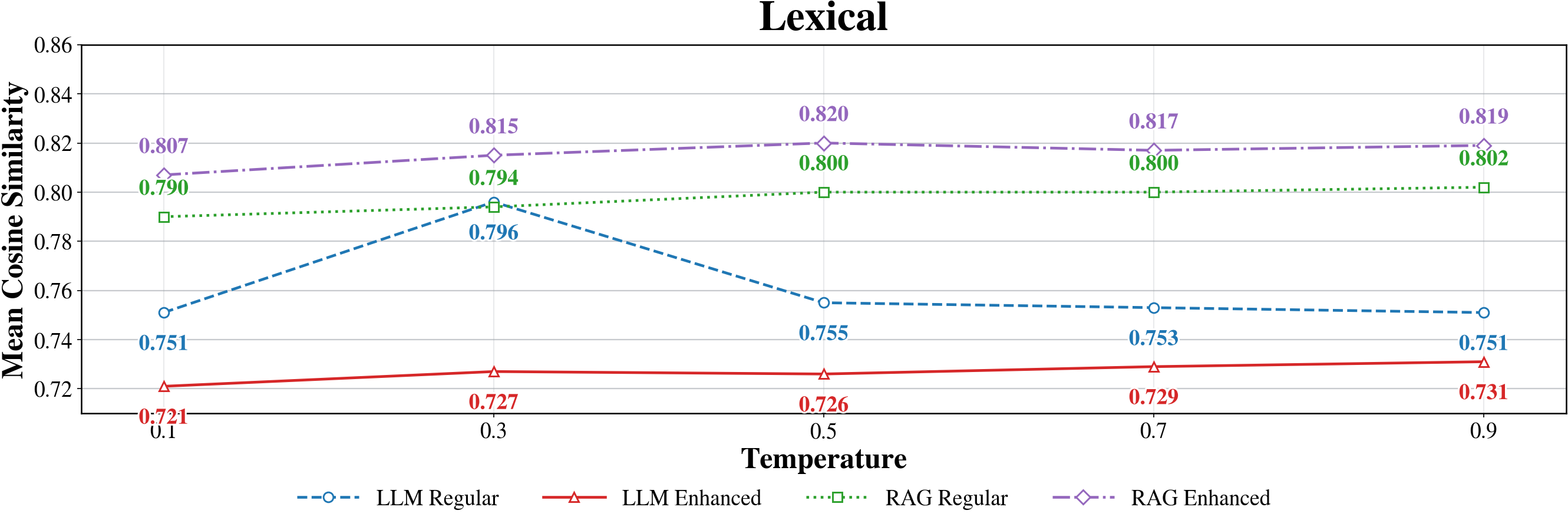}
  \caption[center]{LLM ideological alignment based on lexical representation across different temperatures.}
  \label{fig:linegraph_lexical}
\end{figure}
\begin{figure}[t]
    \centering
\includegraphics[width=0.95\textwidth]{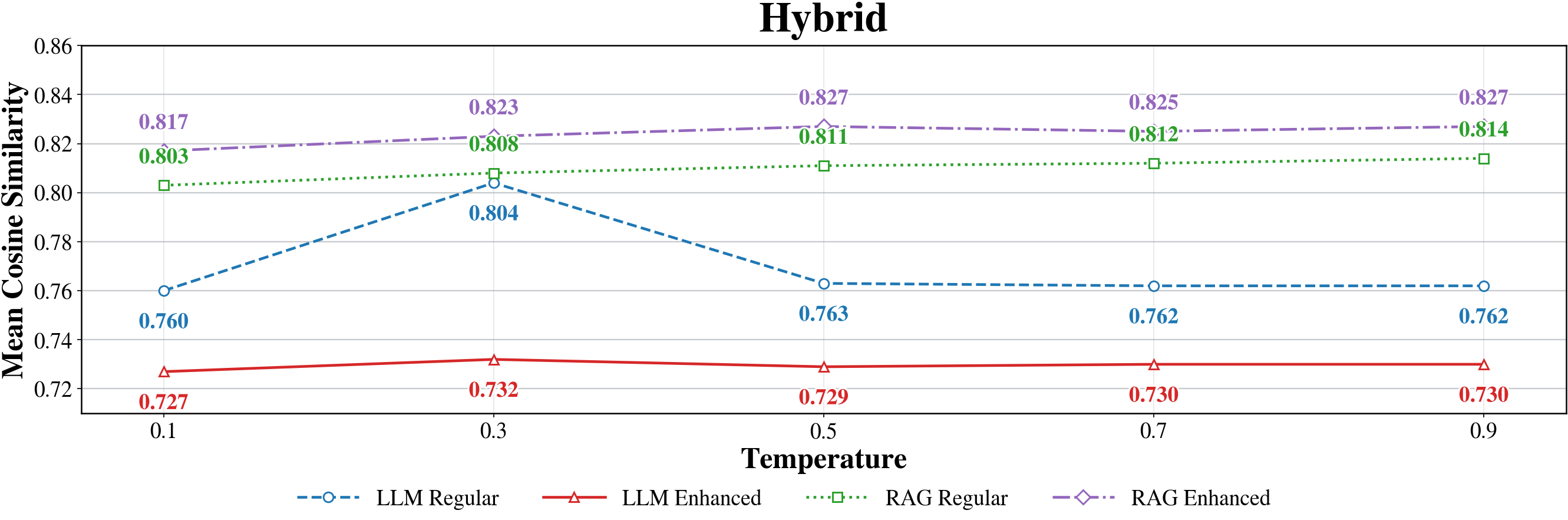}
  \caption[center]{LLM ideological alignment based on the hybrid representation across different temperatures.}
  \label{fig:linegraph_hybrid}
\end{figure}

\section{Main Results}

Figures~\ref{fig:linegraph_semantic}--\ref{fig:linegraph_hybrid} compare discourse alignment across four generation conditions over five sampling temperatures. This setup allows us to assess how sampling temperature interacts with retrieval grounding under regular and enhanced prompting. Across all temperatures and in all three evaluation text representations, the RAG-based configurations achieve higher alignment compared to the LLM-only generation. This pattern is encountered in both prompt strategies, with RAG combined with enhanced prompting providing the best alignment with the reference ideological texts. This suggests that ideological cues in the prompt are most effective when reinforced by the retrieved ideological context. Note that temperature influences the variability of alignment differently across conditions, revealing distinct sensitivity to sampling randomness depending on whether retrieval is available.

For instance, RAG with enhanced prompting yields comparatively lower alignment at the most deterministic setting, i.e., temperature 0.1, increasing to a maximum at the moderate temperature of 0.5, with alignment becoming stable as temperature increases. RAG with regular prompting shows a more gradual and monotonic improvement as temperature values increase, reaching its highest alignment when the temperature is set to 0.9. This indicates that when retrieval grounding is present, higher sampling diversity can be exploited without substantially degrading alignment, likely because the retrieved evidence already constrains generation.

In contrast, LLM-only generation is consistently less aligned than retrieval-grounded generation. With enhanced prompting, alignment remains the lowest among all configurations, exhibiting only a small increase around temperature 0.3, followed by a drop and then near-stability. This suggests that adding discourse-oriented prompt enhancements alone cannot improve alignment, and additional randomness does not compensate for the lack of grounding. With regular prompting, alignment increases sharply from temperature 0.1 to 0.3 but then declines and stabilizes at higher temperatures, producing a clear rise-and-fall pattern. This behavior implies that modest stochasticity may temporarily improve ideological alignment, whereas larger temperatures introduce variability that the model cannot adequately control without retrieval support, making this configuration more sensitive to temperature changes overall.

Overall, the results suggest that temperature affects response discourse alignment differently depending on the generation setup. RAG-based configurations tend to slightly increase alignment as temperature and randomness increase. LLM-only models, on the other hand, show more sensitivity to temperature. 

\subsection{Model-Dependent Alignment}

\begin{figure}[t]
\includegraphics[width=\textwidth]{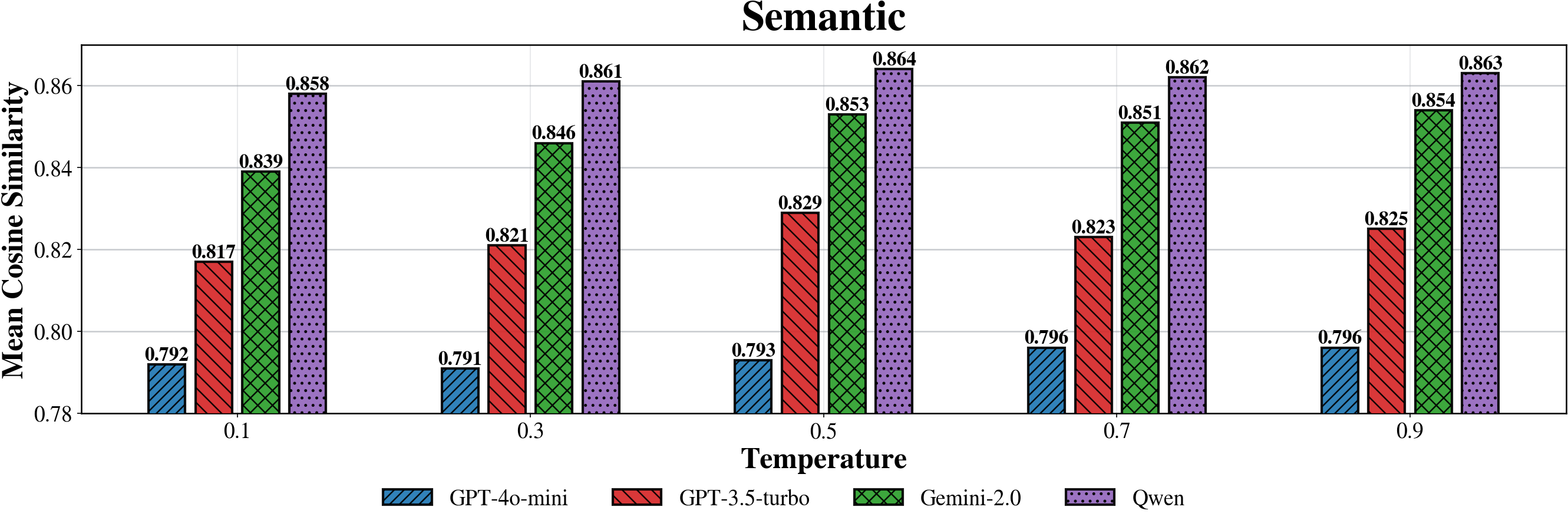}
    \caption{RAG-based ideological alignment based on semantic representation across different temperatures.}
    \label{fig:model-sementic-line}
\end{figure}
\begin{figure}[t]
\includegraphics[width=\textwidth]{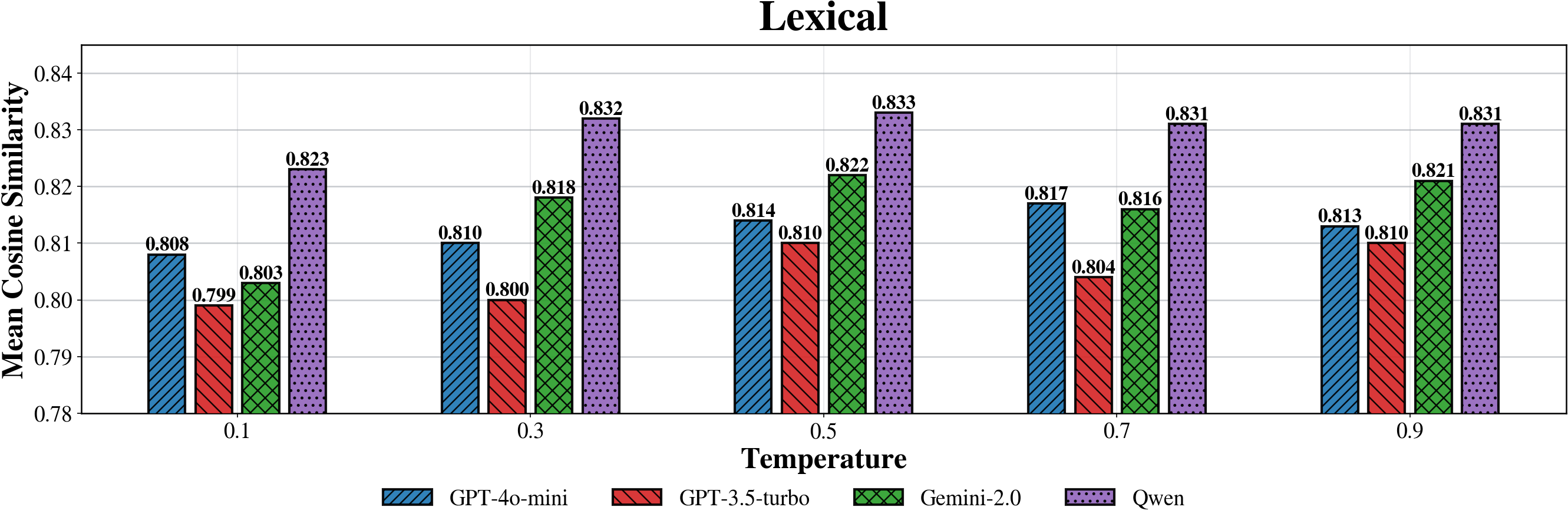}
         \caption{RAG-based ideological alignment based on lexical representation across different temperatures.}
          \label{fig:model-lexical-line}
\end{figure}
\begin{figure}
\includegraphics[width=\textwidth]{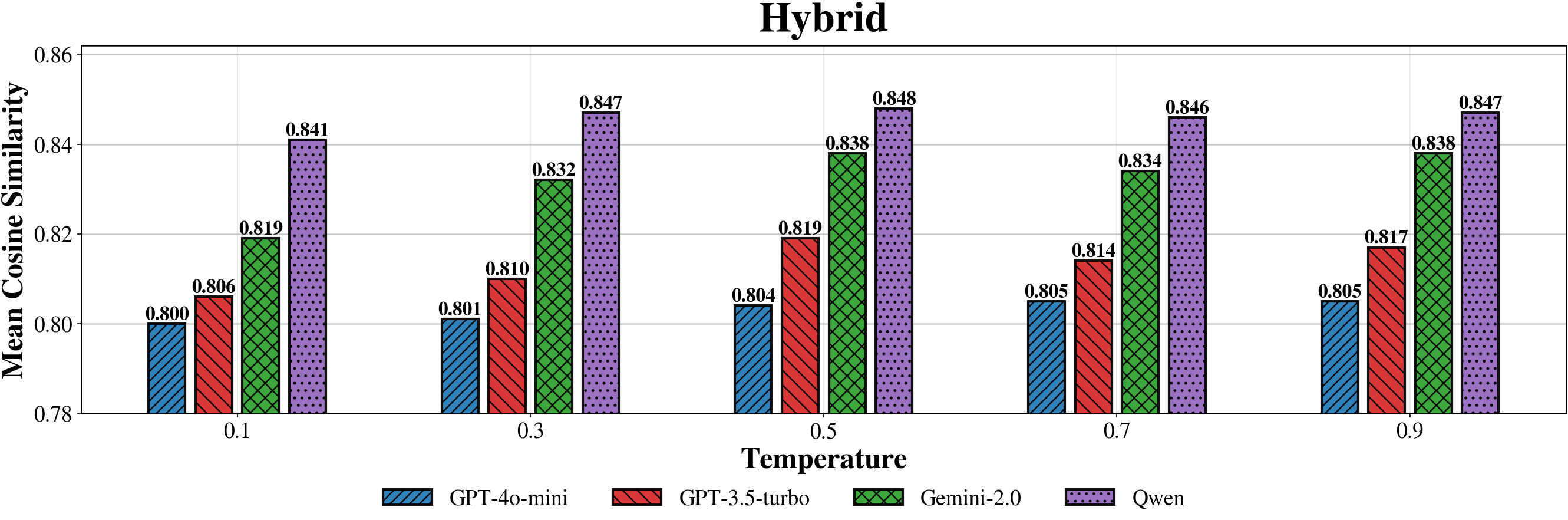}
   \caption{RAG-based ideological alignment based on the hybrid representation across different temperatures.}
    \label{fig:model-hybrid-line}
\end{figure}

Figures~\ref{fig:model-sementic-line}-\ref{fig:model-hybrid-line} present results in terms of mean cosine similarities for semantic, lexical, and hybrid metrics, respectively. We consider four different LLM engines, i.e., GPT-4o-mini, GPT-3.5-turbo, Gemini-2.0 and Qwen, based on the RAG pipeline with enhanced prompting, which was selected as it produced the highest similarity to the ideological reference texts in the preceding experiment. These configurations are tested across five different temperatures. This setting allows for the analysis to isolate how model choice and temperature affect discourse alignment without additional variability from alternative prompting or generation setups. In semantic similarity (see Figure~\ref{fig:model-sementic-line}), alignment differs between models but remains stable across temperatures, with higher temperatures providing better alignment overall. Qwen2.5 achieves the highest alignment, providing similarity scores above 0.85 with its peak at 0.864 for $T=0.5$. It is followed by Gemini-2.0 with scores between 0.84 and 0.85. GPT-3.5-turbo offers moderate similarities and GPT-4o-mini provides the lowest score. Across models, semantic alignment is generally weakest at the most deterministic setting, i.e., $T=0.1$, and increases slightly as temperature rises. Lexical similarity (see Figure~\ref{fig:model-lexical-line}) shows greater temperature sensitivity than semantic similarity, indicating that word-level overlap is more affected by changes on sampling temperature. Qwen2.5 still leads at all temperatures, remaining above 0.82 and peaking at 0.833 at $T=0.5$, followed by Gemini-2.0 and GPT-4o-mini, with GPT-3.5-turbo providing the lowest scores. Note that the Lexical alignment is typically lowest at $T=0.1$ and varies more across the temperature range. For the hybrid similarity (see Figure~\ref{fig:model-hybrid-line}), the pattern falls between the two previous representations, being more stable than lexical alignment but slightly more sensitive to the temperature compared to the semantic alignment. Qwen2.5 still provides the highest scores, followed by Gemini-2.0 and GPT-3.5-turbo, while GPT-4o-mini remains the lowest yet exhibits the least variation across temperatures. These results show consistent cross-model ranking across all three metrics. Additionally, models like Qwen seem more susceptible to ideological influence. Results also suggest that larger models, such as GPT-4o-mini, capable of encoding more world knowledge, are less influenced by ideological texts present in the external knowledge.

\subsection{Statistical Analysis of Temperature and Prompt Type}
Table~\ref{tab:anova_summary} presents the effects of temperature and prompt on discourse alignment based on an ANOVA test. According to the results, temperature has a statistically significant effect on all three similarity metrics, achieving \textit{p}~=0.00018,  \textit{p}=0.00318 and \textit{p}=0.00062, respectively, for the experiments based on semantic, lexical and hybrid representations. This indicates a moderate to strong influence on discourse alignment across decoding conditions. The hybrid metric shows the largest effect size ($\eta^2$=0.244), suggesting that combined semantic–lexical coherence is the most sensitive to temperature changes. Prompt type is also statistically significant across all metrics (\textit{p}=0.00855, $\eta^2$=0.637 for semantic; \textit{p}=0.00504, $\eta^2$=0.571 for lexical; and \textit{p}=0.00349, $\eta^2$=~0.698 for hybrid), with large effect sizes ($\eta^2 > 0.5$) that indicate prompting strategy strongly shapes discourse alignment. In particular, enhanced prompts containing ideological-dimension cues consistently yield higher cosine similarity than regular prompts, and the greatest improvement appears in the hybrid metric, implying that these cues simultaneously strengthen conceptual and lexical coherence.

\begin{table}[t]
\centering
\rowcolors{0}{}{}
\caption{Repeated-measures ANOVA results for Temperature and Prompt Type effects across Semantic, Lexical, and Hybrid metrics.}
\begin{tabular}{lcccccc}
\toprule
Metric & Effect & F & df & p-value & $\eta^2$ \\
\midrule
\multirow{2}{*}{Semantic} 
 & Temperature & 13.998 & (4,12) & 0.00018 & 0.191 \\
 & Prompt Type & 38.140 & (1,3) & 0.00855 & 0.637 \\
    \hdashline
\multirow{2}{*}{Lexical} 
 & Temperature & 7.314 & (4,12) & 0.00318 & 0.162 \\
 & Prompt Type & 55.283 & (1,3) & 0.00504 & 0.571 \\
    \hdashline

\multirow{2}{*}{Hybrid} 
 & Temperature & 10.746 & (4,12) & 0.00062 & 0.244 \\
 & Prompt Type & 71.247 & (1,3) & 0.00349 & 0.698 \\
\bottomrule
\end{tabular}
\\[1ex]
\textit{Note.} $\eta^2$ = generalized eta-squared; n.s. = not significant; † = marginal ($p<.10$); * = $p<.05$; ** = $p<.01$; *** = $p<.001$.
\label{tab:anova_summary}
\end{table}

\subsection{Lexical Diversity, Entropy, and Output Length}
\begin{figure}[t]
\includegraphics[width=\textwidth]{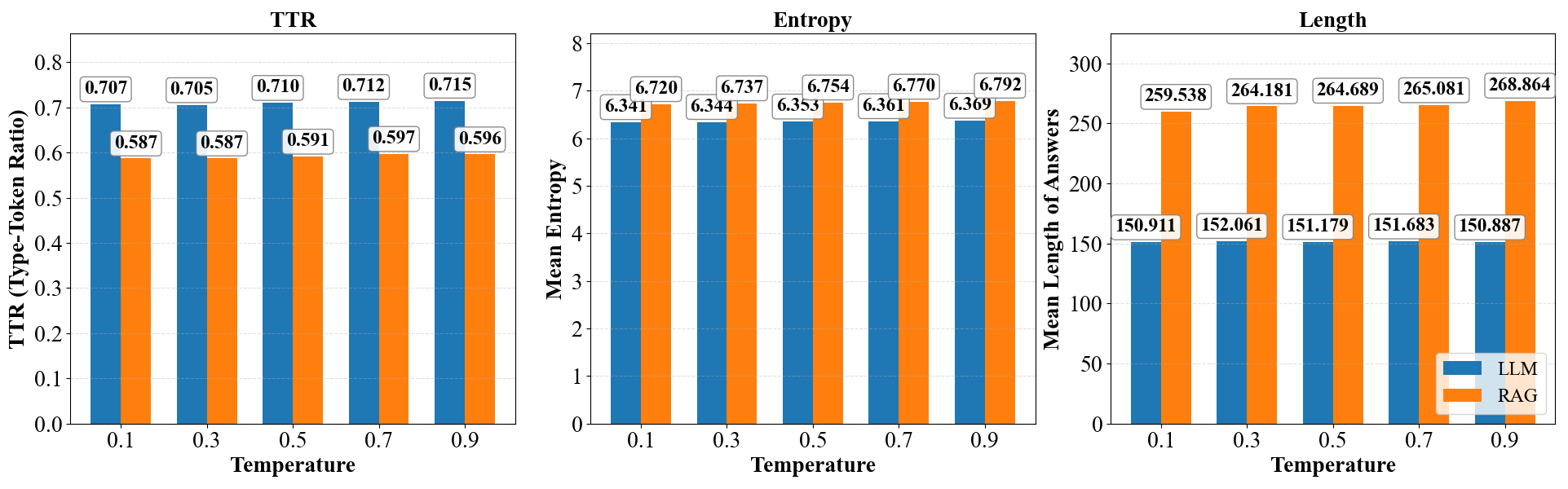}
   \caption{Temperature-Dependent Variation in TTR, Entropy, and Length under Enhanced Prompting}
    \label{fig:TTR}
\end{figure}

Figure~\ref{fig:TTR} presents a comparative analysis of Lexical diversity (TTR), mean entropy, and output length across all temperatures for the LLM-only and RAG settings under the Enhanced prompting. The results show that the LLM-only setting consistently produces outputs with higher lexical diversity than the RAG across all temperatures, with only minor variation as temperature increases. This suggests that while the retrieval pipeline limits the TTR for generated responses, the temperature itself does not significantly influence lexical diversity. The mean entropy shows a consistent and modest rise across temperatures, increasing from 6.341 at $T=0.1$ to 6.369 at $T=0.9$ for the LLM, and from 6.720 at $T=0.1$ to 6.792 at $T=0.9$ for the RAG, suggesting that higher temperatures induce greater randomness in word selection for both LLM-only and RAG systems. The mean length of answers remains relatively stable across temperatures for both settings. However, the RAG consistently produces longer responses than the LLM. For the LLM, response length fluctuates only slightly, ranging from 150.9 to 152.1 tokens, indicating minimal sensitivity to temperature. In contrast, the RAG’s output length increases gradually from 259.5 at $T=0.1$ to 268.9 at $T=0.9$, suggesting that higher sampling temperatures encourage the RAG system to generate marginally longer and more elaborated outputs.  

\begin{figure}
\includegraphics[width=0.9\textwidth]{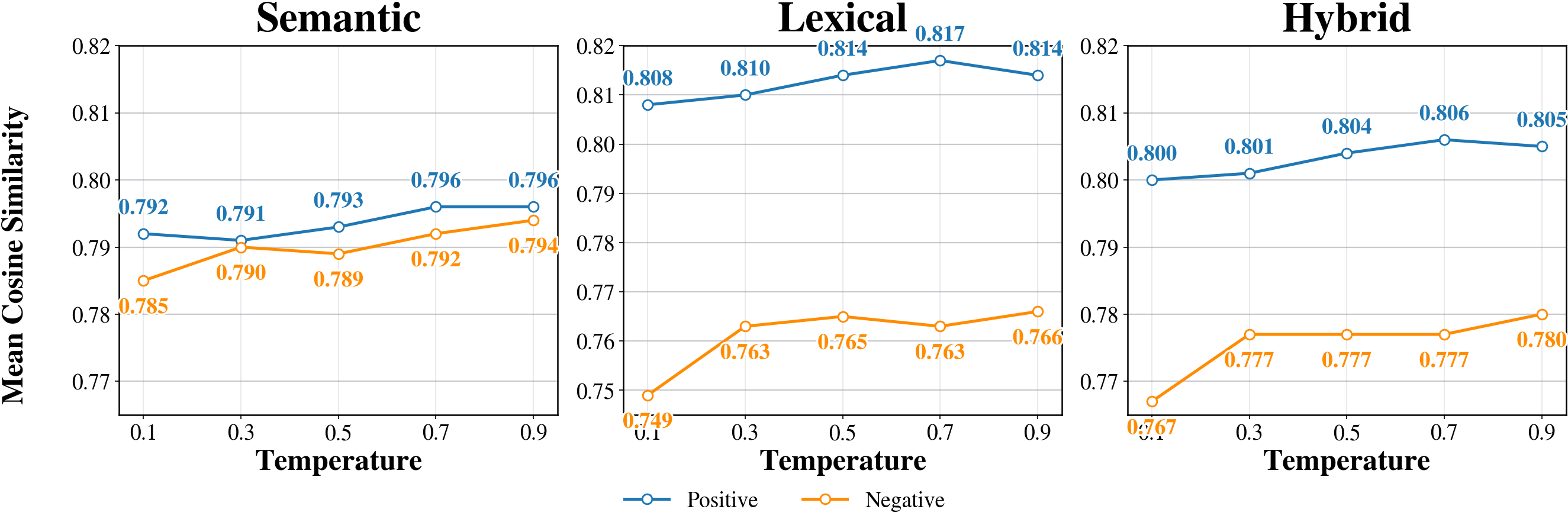}
   \caption{Idelogical discourse alignment for GPT-4o-mini and negative prompting.}
    \label{fig:negative}
\end{figure}

\subsection{Negative Prompt Alignment}
Figure~\ref{fig:negative} presents the results for the experiments with negative prompts. We can observe that compared to the regular settings, the negative prompt consistently lowered discourse alignment. Among the three metrics, the lexical exhibited the largest overall reduction, suggesting that lexical choices are more sensitive to prompt polarity. Additionally, the LMDA is grounded on lexical analysis, which may also explain why the lexical representation better captures ideological variability. In contrast, the semantic metric showed the smallest reduction, indicating that core semantic structures are more resistant to changes in prompt polarity. For both promptings, alignment is lowest at the lowest temperature. Nevertheless, the temperatures yielding the highest alignment differ: positive prompting reaches its maximum at a moderate temperature, i.e., $T=0.7$, whereas negative prompting peaks at $ T=0.9$. Overall, alignment improves as temperature increases.

\section{Limitations}
We acknowledge that this study has some limitations. First, our findings are related to COVID-19 treatments administered between 2020 and 2022. The treatment content during that period was characterized by acute public health urgency, scientific uncertainty, and intense political polarization, all of which could have affected the findings. Second, while this study is a valuable case study for exploring ideological discourses in high-stakes settings, it remains unclear whether the patterns we identify can be generalized to non-medical domains, such as climate change contexts. Third, our evaluations rely on semantic, lexical, and hybrid similarity measures, which may be insufficient for assessing ideological discourse alignment, and human evaluation may be needed to corroborate the findings. Future work may further test the robustness of this approach across diverse ideological debates, determine which findings reflect general mechanisms of ideological encoding and which are specific to pandemic discourse.

\section{Conclusion}
This study investigates the effect of sampling temperature on the transfer of ideological discourse in Retrieval Augmented Generation (RAG) systems across four large language models, different prompting, and system configurations. We compare LLM-only and RAG pipelines using both regular prompts and enhanced prompts enriched with ideological metadata. Model outputs are evaluated using semantic, lexical, and hybrid cosine similarity measures against reference texts to assess ideological discourse alignment across multiple temperature settings.

Temperature significantly impacts alignment across all metrics. However, its effect strongly depends on the system configuration. LLM-only systems exhibit greater variability across temperature, suggesting that stochastic decoding more strongly affects ideological positioning in the absence of external grounding. In contrast, in the RAG setting, ideological alignment remains comparatively stable across temperatures, suggesting that retrieval grounding constrains temperature-driven ideological drift.

Overall, while sampling temperature contributes to ideological transfer, its effect is structurally moderated by retrieval and prompting design. Ideological transfer in grounded systems arises from the interaction between decoding stochasticity, metadata-based prompting, and retrieval constraints, rather than temperature alone.

\section*{Competing Interests}
The authors declare no competing interests.
\section*{Use of AI}
The authors used artificial intelligence (AI) tools, ChatGPT and Grammarly, to improve the clarity and readability of the text. All content was reviewed and verified by the authors, who take full responsibility for the final manuscript.

\printbibliography

\appendix
\section{Supplementary Materials}
\label{App}
This appendix contains supplementary tables and extended results referenced in the main text. These materials offer additional detail on the temperature-wise analyses and discourse alignment metrics that complement the findings reported in the Results section. 


\begin{table*}[!t]
\centering
\scriptsize
\rowcolors{0}{}{}
\setlength{\tabcolsep}{8pt}
\renewcommand{\arraystretch}{1.15}

\caption{Top lexical items and loadings for Dimensions~1--3. Higher loadings indicate stronger association with a pole.}
\label{tab:typical_vocabularies}

\begin{tabularx}{\textwidth}{@{}>{\raggedright\arraybackslash}X >{\raggedright\arraybackslash}X@{}}

\multicolumn{2}{c}{\textbf{Dimension 1}}\\
\toprule
\textbf{Positive pole (Controversial)} & \textbf{Negative pole (Endorsed)}\\
\midrule
ventilation + mechanical \hfill (0.84) & media + social \hfill (0.89)\\
discharged + patients \hfill (0.80) & learning + online \hfill (0.83)\\
hospitalized + patients \hfill (0.79) & future + limitations \hfill (0.82)\\
intensive + admission \hfill (0.78) & education + online \hfill (0.82)\\
respiratory + rate \hfill (0.76) & anxiety + symptoms \hfill (0.78)\\
treatment + patients \hfill (0.74) & completed + participants \hfill (0.76)\\
clinical + patients \hfill (0.73) & directions + limitations \hfill (0.75)\\
received + patients \hfill (0.71) & students + health \hfill (0.75)\\
placebo + patients \hfill (0.71) & anxiety + depression \hfill (0.75)\\
\bottomrule

\\[-2pt]
\multicolumn{2}{c}{\textbf{Dimension 2}}\\
\toprule
\textbf{Positive pole (Endorsed)} & \textbf{Negative pole (Controversial)}\\
\midrule
translation + omissions \hfill (1.40) & protein + creative \hfill (0.53)\\
material + content \hfill (1.40) & hydroxychloroquine + azithromycin \hfill (0.50)\\
reliability + warrant \hfill (1.40) & respiratory + rate \hfill (0.47)\\
placed + arising \hfill (1.40) & received + patients \hfill (0.46)\\
adaptation + arising \hfill (1.40) & treatment + compared \hfill (0.45)\\
arising + omissions \hfill (1.40) & controlled + randomized \hfill (0.45)\\
content + reliance \hfill (1.40) & combination + alone \hfill (0.44)\\
content + supplemental \hfill (1.30) & syndrome + acute \hfill (0.44)\\
material + supplemental \hfill (1.28) & trial + randomized \hfill (0.43)\\
peer + provenance \hfill (1.27) & azithromycin + combination \hfill (0.42)\\
\bottomrule

\\[-2pt]
\multicolumn{2}{c}{\textbf{Dimension 3}}\\
\toprule
\textbf{Positive pole (Controversial)} & \textbf{Negative pole (Endorsed)}\\
\midrule
interval + prolongation \hfill (0.99) & contains + information \hfill (0.91)\\
hydroxychloroquine + alone \hfill (0.97) & online + supplementary \hfill (0.89)\\
rheumatoid + systemic \hfill (0.92) & material + online \hfill (0.87)\\
predictors + independent \hfill (0.90) & information + supplementary \hfill (0.85)\\
survival + analyses \hfill (0.75) & reviews + reporting \hfill (0.77)\\
treatment + outcomes \hfill (0.73) & materials + availability \hfill (0.74)\\
steroid + therapy \hfill (0.69) & applicable + consent \hfill (0.72)\\
twice + followed \hfill (0.68) & review + included \hfill (0.69)\\
alone + treatment \hfill (0.66) & declarations + ethics \hfill (0.69)\\
adjusted + mortality \hfill (0.63) & participate + approval \hfill (0.65)\\
\bottomrule

\end{tabularx}
\end{table*}


\begin{table*}[!htb]
\centering
\rowcolors{0}{}{}
\small
\caption{Questions for Endorsed Treatments (ET) vs. Controversial Treatments (CT)}
\label{table:topic}
\setlength{\tabcolsep}{6pt}
\renewcommand{\arraystretch}{1.2}
\begin{tabularx}{\textwidth}{@{}XX@{}}
\toprule
\multicolumn{1}{c}{\textbf{Endorsed Treatments (ET)}} & \multicolumn{1}{c}{\textbf{Controversial Treatments (CT)}} \\
\midrule

\multicolumn{2}{c}{\textbf{Dimension 1}} \\
\midrule

\textit{Mental Health Impacts} (--)
\newline \textbf{Q1:} How has the pandemic affected mental health in the U.S.?
\newline \textbf{Q2:} How can I cope with anxiety and stress?

\vspace{0.5em}
\textit{Role of Social Media in Mental Health} (--)
\newline \textbf{Q1:} Does social media worsen mental health issues?
\newline \textbf{Q2:} How can social media improve mental health?

\vspace{0.5em}
\textit{Coping Mechanisms During the Pandemic} (--)
\newline \textbf{Q1:} What are the best ways to cope with stress?
\newline \textbf{Q2:} How can I stay mentally healthy at home? 

& \textit{Hydroxychloroquine and Azithromycin Use} (+)
\newline \textbf{Q1:} Is hydroxychloroquine effective when combined with azithromycin?
\newline \textbf{Q2:} What are the risks and side effects?

\vspace{0.5em}
\textit{In-Hospital Mortality} (+)
\newline \textbf{Q1:} What steps is the government taking for critically ill patients?
\newline \textbf{Q2:} How do hospitals reduce mortality rates?

\vspace{0.5em}
\textit{Risk Factors and Predictors of Mortality} (+)
\newline \textbf{Q1:} How do doctors identify highest-risk patients?
\newline \textbf{Q2:} Who is most at risk of dying from COVID-19? \\

\midrule
\multicolumn{2}{c}{\textbf{Dimension 2}} \\
\midrule

\textit{Global and Regional Responses} (+)
\newline \textbf{Q1:} How have different countries handled the pandemic?
\newline \textbf{Q2:} What has the U.S. done compared to other regions?

\vspace{0.5em}
\textit{Impact on Health Systems and Vulnerable Groups} (+)
\newline \textbf{Q1:} How has COVID-19 affected hospitals and workers?
\newline \textbf{Q2:} Why are certain groups more affected?

\vspace{0.5em}
\textit{Innovative Approaches to Treatment} (+)
\newline \textbf{Q1:} What are the newest treatments being developed?
\newline \textbf{Q2:} How are hospitals using technology?

& \textit{Hydroxychloroquine Use in COVID-19} (--)
\newline \textbf{Q1:} Is hydroxychloroquine still being used?
\newline \textbf{Q2:} What are the risks of taking it?

\vspace{0.5em}
\textit{Risk Factors and Predictors of Mortality} (--)
\newline \textbf{Q1:} What conditions increase the risk of dying?
\newline \textbf{Q2:} Who is most likely to die, and why?

\vspace{0.5em}
\textit{Clinical and Epidemiological Characteristics} (--)
\newline \textbf{Q1:} How does COVID-19 affect people by age/conditions?
\newline \textbf{Q2:} What are typical profiles of severe patients? \\

\midrule
\multicolumn{2}{c}{\textbf{Dimension 3}} \\
\midrule

\textit{Impact on Public Health Systems} (--)
\newline \textbf{Q1:} How has COVID-19 affected healthcare resources?
\newline \textbf{Q2:} What policies help healthcare workers?

\vspace{0.5em}
\textit{Efficacy and Safety of Interventions} (--)
\newline \textbf{Q1:} How safe are available treatments?
\newline \textbf{Q2:} Do treatments help patients recover faster?

\vspace{0.5em}
\textit{Risk Assessment and Management} (--)
\newline \textbf{Q1:} How do doctors assess severe illness risk?
\newline \textbf{Q2:} What steps can lower my risk of complications?

& \textit{Hydroxychloroquine as a Treatment Option} (+)
\newline \textbf{Q1:} Does hydroxychloroquine actually work?
\newline \textbf{Q2:} What are the side effects?

\vspace{0.5em}
\textit{Effectiveness and Safety of Treatments} (+)
\newline \textbf{Q1:} Are treatments like remdesivir safe?
\newline \textbf{Q2:} How do treatments improve patient outcomes?

\vspace{0.5em}
\textit{Impact on Mortality and Viral Clearance} (+)
\newline \textbf{Q1:} How do doctors assess severe illness risk?
\newline \textbf{Q2:} How does the virus cause death in severe cases? \\

\bottomrule
\end{tabularx}
\end{table*}

\begin{table}
\caption{Three Dimensions of COVID-19 Treatment Discourse}
\label{tab:dimensions}
\centering
\rowcolors{0}{}{}
\small
\begin{tabularx}{\textwidth}{@{}c>{\hsize=1.0\hsize}X>{\hsize=1.0\hsize}X@{}}
\toprule
\textbf{Dim} & \textbf{Positive Pole} & \textbf{Negative Pole} \\
\midrule

\textbf{1}
&
\textbf{Label:} Endorsement of controversial medications (hydroxychloroquine) as effective treatments, promoting pharmaceutical messianism.

\textbf{Description:} Texts adopt promotional stances toward controversial treatments, using rhetorical strategies to create efficacy and urgency despite questionable scientific backing.

\textbf{Bias:} Controversial.
&
\textbf{Label:} Examination of psychological impacts from the pandemic, social distancing, and public health measures.

\textbf{Description:} Research investigates mental health consequences (anxiety, depression, stress, isolation) using empirical language to highlight psychological distress and social implications.

\textbf{Bias:} Endorsed.
\\
\addlinespace

\textbf{2}
&
\textbf{Label:} Adherence to research ethics standards in publication, data availability, liability, and translation.

\textbf{Description:} Texts construct norms around ethical responsibility, transparency, and accountability in COVID-19 treatment research through scientific integrity and trustworthiness.

\textbf{Bias:} Endorsed.
&
\textbf{Label:} Comparative analysis concluding that questionable drugs (hydroxychloroquine, ivermectin) are effective.

\textbf{Description:} Text structure arguments endorsing controversial treatments through linguistic framing, data selection, and rhetorical persuasion despite scientific skepticism.

\textbf{Bias:} Controversial.
\\
\addlinespace

\textbf{3}
&
\textbf{Label:} Promotion of repurposed drugs despite inconclusive evidence, emphasizing lower mortality while downplaying risks.

\textbf{Description:} Texts advocate for widespread use of repurposed drugs (hydroxychloroquine, ivermectin), highlighting potential benefits while minimizing discussions of associated risks.

\textbf{Bias:} Controversial.
&
\textbf{Label:} Critical evaluation of outcomes (mortality, virological eradication, adverse events) with emphasis on open science practices.

\textbf{Description:} Studies prioritize rigorous assessment and transparency, providing access to data, materials, and ethical considerations aligned with open science principles.

\textbf{Bias:} Endorsed.
\\
\bottomrule
\end{tabularx}
\end{table}

\begin{table*}[ht]
\centering
\tiny 
\rowcolors{0}{}{}
\caption{Semantic cosine similarities for Regular and Enhanced prompts across five temperatures (0.1, 0.3, 0.5, 0.7, 0.9)}
\label{tab:semantic_alltemps}
\renewcommand{\arraystretch}{1.15}

\begin{tabular}{|cclcccccccccccc|}
\toprule
& & \textbf{\textbf{\scriptsize Model}} &
\multicolumn{2}{c}{{\scriptsize Dim 1 (+)}} &
\multicolumn{2}{c}{{\scriptsize Dim 1 (-)}} &
\multicolumn{2}{c}{{\scriptsize Dim 2 (+)}} &
\multicolumn{2}{c}{{\scriptsize Dim 2 (-)}} &
\multicolumn{2}{c}{{\scriptsize Dim 3 (+)}} &
\multicolumn{2}{c|}{{\scriptsize Dim 3 (-)}} \\[3pt]
\cmidrule(lr){4-5}
\cmidrule(lr){6-7}
\cmidrule(lr){8-9}
\cmidrule(lr){10-11}
\cmidrule(lr){12-13}
\cmidrule(lr){14-15}
& & & {\scriptsize LLM} & {\scriptsize RAG} & {\scriptsize LLM} & {\scriptsize RAG} & {\scriptsize LLM} & {\scriptsize RAG} & {\scriptsize LLM} & {\scriptsize RAG} &{\scriptsize LLM} & {\scriptsize RAG} & {\scriptsize LLM} & {\scriptsize RAG} \\
\midrule
\multirow{8}{*}{\rotatebox{90}{\textbf{\small   Temperature 0.1}}}
& \multirow{4}{*}{\rotatebox{90}{\textbf{\scriptsize Regular}}}
& {\scriptsize GPT-4o-mini}  & 0.794 & 0.845 & 0.725 & 0.775 & 0.781 & 0.785 & 0.783 & 0.816 & 0.783 & 0.837 & 0.750 & 0.758 \\
& & {\scriptsize GPT-3.5-turbo} & 0.770 & 0.855 & 0.699 & 0.812 & 0.771 & 0.761 & 0.782 & 0.793 & 0.810 & 0.885 & 0.705 & 0.729 \\
& & {\scriptsize Gemini-2.0}    & 0.805 & 0.866 & 0.728 & 0.807 & 0.789 & 0.825 & 0.758 & 0.828 & 0.784 & 0.875 & 0.745 & 0.763 \\
& & {\scriptsize Qwen}          & 0.801 & 0.862 & 0.749 & 0.811 & 0.802 & 0.807 & 0.807 & 0.871 & 0.810 & 0.858 & 0.751 & 0.769 \\
\cmidrule(lr){3-15}
& \multirow{4}{*}{\rotatebox{90}{\textbf{\scriptsize Enhanced}}}
& {\scriptsize GPT-4o-mini}  & 0.751 & 0.804 & 0.732 & 0.812 & 0.588 & 0.628 & 0.738 & 0.847 & 0.797 & 0.851 & 0.756 & 0.807 \\
& & {\scriptsize GPT-3.5-turbo} & 0.763 & 0.847 & 0.757 & 0.840 & 0.666 & 0.724 & 0.732 & 0.810 & 0.782 & 0.891 & 0.734 & 0.791 \\
& & {\scriptsize Gemini-2.0}   & 0.765 & 0.864 & 0.739 & 0.835 & 0.636 & 0.763 & 0.708 & 0.856 & 0.768 & 0.899 & 0.753 & 0.815 \\
& & {\scriptsize Qwen}          & 0.765 & 0.905 & 0.756 & 0.902 & 0.609 & 0.767 & 0.722 & 0.890 & 0.777 & 0.915 & 0.762 & 0.769 \\
\bottomrule
\end{tabular}

\vspace{5pt}

\begin{tabular}{|cclcccccccccccc|}
\toprule
& & \textbf{\textbf{\scriptsize Model}} &
\multicolumn{2}{c}{{\scriptsize Dim 1 (+)}} &
\multicolumn{2}{c}{{\scriptsize Dim 1 (-)}} &
\multicolumn{2}{c}{{\scriptsize Dim 2 (+)}} &
\multicolumn{2}{c}{{\scriptsize Dim 2 (-)}} &
\multicolumn{2}{c}{{\scriptsize Dim 3 (+)}} &
\multicolumn{2}{c|}{{\scriptsize Dim 3 (-)}} \\[3pt]
\cmidrule(lr){4-5}
\cmidrule(lr){6-7}
\cmidrule(lr){8-9}
\cmidrule(lr){10-11}
\cmidrule(lr){12-13}
\cmidrule(lr){14-15}
& & & {\scriptsize LLM} & {\scriptsize RAG} & {\scriptsize LLM} & {\scriptsize RAG} & {\scriptsize LLM} & {\scriptsize RAG} & {\scriptsize LLM} & {\scriptsize RAG} &{\scriptsize LLM} & {\scriptsize RAG} & {\scriptsize LLM} & {\scriptsize RAG} \\
\midrule
\multirow{8}{*}{\rotatebox{90}{\textbf{\small   Temperature 0.3}}}
& \multirow{4}{*}{\rotatebox{90}{\textbf{\scriptsize Regular}}}
& {\scriptsize GPT-4o-mini}  & 0.847 & 0.849 & 0.784 & 0.785 & 0.785 & 0.780 & 0.824 & 0.811 & 0.839 & 0.831 & 0.772 & 0.767 \\
& & {\scriptsize GPT-3.5-turbo} & 0.857 & 0.858 & 0.820 & 0.807 & 0.752 & 0.757 & 0.813 & 0.809 & 0.884 & 0.887 & 0.739 & 0.740 \\
& & {\scriptsize Gemini-2.0}  & 0.805 & 0.867 & 0.725 & 0.781 & 0.793 & 0.782 & 0.756 & 0.798 & 0.786 & 0.874 & 0.746 & 0.757 \\
& & {\scriptsize Qwen}          & 0.901 & 0.902 & 0.899 & 0.898 & 0.766 & 0.777 & 0.909 & 0.894 & 0.919 & 0.922 & 0.780 & 0.777 \\
\cmidrule(lr){3-15}
& \multirow{4}{*}{\rotatebox{90}{\textbf{\scriptsize Enhanced}}}
& {\scriptsize GPT-4o-mini}    & 0.749 & 0.799 & 0.750 & 0.817 & 0.600 & 0.635 & 0.741 & 0.841 & 0.789 & 0.851 & 0.752 & 0.803 \\
& & {\scriptsize GPT-3.5-turbo}& 0.783 & 0.842 & 0.759 & 0.840 & 0.665 & 0.740 & 0.743 & 0.807 & 0.803 & 0.888 & 0.723 & 0.807 \\
& & {\scriptsize Gemini-2.0}  & 0.753 & 0.876 & 0.724 & 0.835 & 0.627 & 0.783 & 0.718 & 0.865 & 0.773 & 0.892 & 0.773 & 0.827 \\
& & {\scriptsize Qwen}        & 0.772 & 0.914 & 0.751 & 0.898 & 0.630 & 0.768 & 0.739 & 0.900 & 0.777 & 0.918 & 0.774 & 0.766 \\
\bottomrule
\end{tabular}

\vspace{5pt}

\begin{tabular}{|cclcccccccccccc|}
\toprule
& & \textbf{\textbf{\scriptsize Model}} &
\multicolumn{2}{c}{{\scriptsize Dim 1 (+)}} &
\multicolumn{2}{c}{{\scriptsize Dim 1 (-)}} &
\multicolumn{2}{c}{{\scriptsize Dim 2 (+)}} &
\multicolumn{2}{c}{{\scriptsize Dim 2 (-)}} &
\multicolumn{2}{c}{{\scriptsize Dim 3 (+)}} &
\multicolumn{2}{c|}{{\scriptsize Dim 3 (-)}} \\[3pt]
\cmidrule(lr){4-5}
\cmidrule(lr){6-7}
\cmidrule(lr){8-9}
\cmidrule(lr){10-11}
\cmidrule(lr){12-13}
\cmidrule(lr){14-15}
& & & {\scriptsize LLM} & {\scriptsize RAG} & {\scriptsize LLM} & {\scriptsize RAG} & {\scriptsize LLM} & {\scriptsize RAG} & {\scriptsize LLM} & {\scriptsize RAG} &{\scriptsize LLM} & {\scriptsize RAG} & {\scriptsize LLM} & {\scriptsize RAG} \\
\midrule
\multirow{8}{*}{\rotatebox{90}{\textbf{\small  Temperature 0.5}}}
& \multirow{4}{*}{\rotatebox{90}{\textbf{\scriptsize Regular}}}
& {\scriptsize GPT-4o-mini}   & 0.795 & 0.839 & 0.725 & 0.775 & 0.772 & 0.782 & 0.783 & 0.808 & 0.789 & 0.832 & 0.748 & 0.759 \\
& & {\scriptsize GPT-3.5-turbo} & 0.767 & 0.851 & 0.713 & 0.816 & 0.765 & 0.760 & 0.788 & 0.832 & 0.799 & 0.895 & 0.723 & 0.734 \\
& &  {\scriptsize Gemini-2.0}   & 0.807 & 0.865 & 0.719 & 0.796 & 0.785 & 0.788 & 0.751 & 0.820 & 0.782 & 0.873 & 0.746 & 0.768 \\
& & {\scriptsize Qwen}         & 0.815 & 0.907 & 0.743 & 0.901 & 0.803 & 0.772 & 0.809 & 0.893 & 0.807 & 0.925 & 0.761 & 0.775 \\
\cmidrule(lr){3-15}
& \multirow{4}{*}{\rotatebox{90}{\textbf{\scriptsize Enhanced}}}
& {\scriptsize GPT-4o-mini}  & 0.744 & 0.795 & 0.738 & 0.818 & 0.608 & 0.641 & 0.740 & 0.846 & 0.788 & 0.848 & 0.755 & 0.809 \\
& & {\scriptsize GPT-3.5-turbo} & 0.780 & 0.862 & 0.749 & 0.845 & 0.639 & 0.747 & 0.747 & 0.826 & 0.798 & 0.891 & 0.723 & 0.801 \\
& & {\scriptsize Gemini-2.0}   & 0.753 & 0.879 & 0.729 & 0.842 & 0.630 & 0.796 & 0.722 & 0.876 & 0.763 & 0.910 & 0.758 & 0.816 \\
& & {\scriptsize Qwen}    & 0.774 & 0.906 & 0.747 & 0.900 & 0.612 & 0.774 & 0.743 & 0.902 & 0.765 & 0.920 & 0.757 & 0.783 \\
\bottomrule
\end{tabular}

\vspace{5pt}

\begin{tabular}{|cclcccccccccccc|}
\toprule
& & \textbf{\textbf{\scriptsize Model}} &
\multicolumn{2}{c}{{\scriptsize Dim 1 (+)}} &
\multicolumn{2}{c}{{\scriptsize Dim 1 (-)}} &
\multicolumn{2}{c}{{\scriptsize Dim 2 (+)}} &
\multicolumn{2}{c}{{\scriptsize Dim 2 (-)}} &
\multicolumn{2}{c}{{\scriptsize Dim 3 (+)}} &
\multicolumn{2}{c|}{{\scriptsize Dim 3 (-)}} \\[3pt]
\cmidrule(lr){4-5}
\cmidrule(lr){6-7}
\cmidrule(lr){8-9}
\cmidrule(lr){10-11}
\cmidrule(lr){12-13}
\cmidrule(lr){14-15}
& & & {\scriptsize LLM} & {\scriptsize RAG} & {\scriptsize LLM} & {\scriptsize RAG} & {\scriptsize LLM} & {\scriptsize RAG} & {\scriptsize LLM} & {\scriptsize RAG} &{\scriptsize LLM} & {\scriptsize RAG} & {\scriptsize LLM} & {\scriptsize RAG} \\
\midrule
\multirow{8}{*}{\rotatebox{90}{\textbf{\small Temperature 0.7}}}
& \multirow{4}{*}{\rotatebox{90}{\textbf{\scriptsize Regular}}}
& {\scriptsize GPT-4o-mini}   & 0.801 & 0.845 & 0.718 & 0.785 & 0.777 & 0.779 & 0.787 & 0.825 & 0.797 & 0.838 & 0.748 & 0.755 \\
& & {\scriptsize GPT-3.5-turbo} & 0.773 & 0.850 & 0.722 & 0.824 & 0.786 & 0.776 & 0.774 & 0.824 & 0.799 & 0.877 & 0.725 & 0.721 \\
& & {\scriptsize Gemini-2.0}   & 0.804 & 0.859 & 0.717 & 0.794 & 0.792 & 0.790 & 0.756 & 0.815 & 0.776 & 0.887 & 0.747 & 0.756 \\
& & {\scriptsize Qwen}        & 0.810 & 0.898 & 0.736 & 0.898 & 0.787 & 0.773 & 0.818 & 0.911 & 0.806 & 0.928 & 0.761 & 0.768 \\
\cmidrule(lr){3-15}
& \multirow{4}{*}{\rotatebox{90}{\textbf{\scriptsize Enhanced}}}
& {\scriptsize GPT-4o-mini}  & 0.746 & 0.811 & 0.746 & 0.808 & 0.597 & 0.652 & 0.744 & 0.847 & 0.784 & 0.856 & 0.755 & 0.803 \\
& & {\scriptsize GPT-3.5-turbo} & 0.758 & 0.848 & 0.741 & 0.842 & 0.628 & 0.730 & 0.739 & 0.826 & 0.801 & 0.895 & 0.727 & 0.795 \\
& & {\scriptsize Gemini-2.0}   & 0.756 & 0.878 & 0.723 & 0.832 & 0.625 & 0.792 & 0.713 & 0.859 & 0.763 & 0.914 & 0.762 & 0.829 \\
& & {\scriptsize Qwen}        & 0.770 & 0.909 & 0.754 & 0.900 & 0.624 & 0.767 & 0.739 & 0.901 & 0.785 & 0.927 & 0.764 & 0.767 \\
\bottomrule
\end{tabular}

\vspace{5pt}

\begin{tabular}{|cclcccccccccccc|}
\toprule
& & \textbf{\textbf{\scriptsize Model}} &
\multicolumn{2}{c}{{\scriptsize Dim 1 (+)}} &
\multicolumn{2}{c}{{\scriptsize Dim 1 (-)}} &
\multicolumn{2}{c}{{\scriptsize Dim 2 (+)}} &
\multicolumn{2}{c}{{\scriptsize Dim 2 (-)}} &
\multicolumn{2}{c}{{\scriptsize Dim 3 (+)}} &
\multicolumn{2}{c|}{{\scriptsize Dim 3 (-)}} \\[3pt]
\cmidrule(lr){4-5}
\cmidrule(lr){6-7}
\cmidrule(lr){8-9}
\cmidrule(lr){10-11}
\cmidrule(lr){12-13}
\cmidrule(lr){14-15}
& & & {\scriptsize LLM} & {\scriptsize RAG} & {\scriptsize LLM} & {\scriptsize RAG} & {\scriptsize LLM} & {\scriptsize RAG} & {\scriptsize LLM} & {\scriptsize RAG} &{\scriptsize LLM} & {\scriptsize RAG} & {\scriptsize LLM} & {\scriptsize RAG} \\
\midrule
\multirow{8}{*}{\rotatebox{90}{\textbf{\small Temperature 0.9}}}
& \multirow{4}{*}{\rotatebox{90}{\textbf{\scriptsize Regular}}}
& {\scriptsize GPT-4o-mini}   & 0.801 & 0.846 & 0.726 & 0.788 & 0.778 & 0.773 & 0.787 & 0.831 & 0.797 & 0.839 & 0.755 & 0.771 \\
& & {\scriptsize GPT-3.5-turbo} & 0.779 & 0.852 & 0.713 & 0.826 & 0.777 & 0.776 & 0.782 & 0.826 & 0.796 & 0.887 & 0.726 & 0.733 \\
& & {\scriptsize Gemini-2.0}   & 0.807 & 0.860 & 0.712 & 0.807 & 0.782 & 0.807 & 0.755 & 0.791 & 0.783 & 0.869 & 0.738 & 0.752 \\
& & {\scriptsize Qwen}          & 0.816 & 0.906 & 0.757 & 0.899 & 0.786 & 0.768 & 0.807 & 0.895 & 0.813 & 0.932 & 0.759 & 0.776 \\
\cmidrule(lr){3-15}
& \multirow{4}{*}{\rotatebox{90}{\textbf{\scriptsize Enhanced}}}
& {\scriptsize GPT-4o-mini}   & 0.750 & 0.805 & 0.732 & 0.810 & 0.601 & 0.655 & 0.745 & 0.840 & 0.793 & 0.853 & 0.753 & 0.812 \\
& & {\scriptsize GPT-3.5-turbo} & 0.786 & 0.840 & 0.737 & 0.852 & 0.628 & 0.745 & 0.750 & 0.815 & 0.798 & 0.899 & 0.736 & 0.799 \\
& & {\scriptsize Gemini-2.0}   & 0.755 & 0.878 & 0.721 & 0.835 & 0.611 & 0.809 & 0.701 & 0.874 & 0.747 & 0.904 & 0.770 & 0.823 \\
& & {\scriptsize Qwen}   & 0.780 & 0.907 & 0.747 & 0.901 & 0.615 & 0.770 & 0.736 & 0.897 & 0.771 & 0.927 & 0.766 & 0.777 \\
\bottomrule
\end{tabular}
\end{table*}
\begin{table*}[ht]
\centering
\tiny
\rowcolors{0}{}{}
\caption{Lexical cosine similarities for Regular and Enhanced prompts across five temperatures}
\label{tab:lexical_alltemps}
\renewcommand{\arraystretch}{1.15}

\begin{tabular}{|cclcccccccccccc|}
\toprule
& & \textbf{\textbf{\scriptsize Model}} &
\multicolumn{2}{c}{{\scriptsize Dim 1 (+)}} &
\multicolumn{2}{c}{{\scriptsize Dim 1 (-)}} &
\multicolumn{2}{c}{{\scriptsize Dim 2 (+)}} &
\multicolumn{2}{c}{{\scriptsize Dim 2 (-)}} &
\multicolumn{2}{c}{{\scriptsize Dim 3 (+)}} &
\multicolumn{2}{c|}{{\scriptsize Dim 3 (-)}} \\[3pt]
\cmidrule(lr){4-5}
\cmidrule(lr){6-7}
\cmidrule(lr){8-9}
\cmidrule(lr){10-11}
\cmidrule(lr){12-13}
\cmidrule(lr){14-15}
& & & {\scriptsize LLM} & {\scriptsize RAG} & {\scriptsize LLM} & {\scriptsize RAG} & {\scriptsize LLM} & {\scriptsize RAG} & {\scriptsize LLM} & {\scriptsize RAG} &{\scriptsize LLM} & {\scriptsize RAG} & {\scriptsize LLM} & {\scriptsize RAG} \\
\midrule
\multirow{8}{*}{\rotatebox{90}{\textbf{\small Temperature 0.1}}}
& \multirow{4}{*}{\rotatebox{90}{\textbf{\scriptsize Regular}}}
& {\scriptsize GPT-4o-mini}  & 0.669 & 0.708 & 0.679 & 0.699 & 0.885 & 0.887 & 0.779 & 0.830 & 0.754 & 0.799 & 0.846 & 0.848 \\
& & {\scriptsize GPT-3.5-turbo}  & 0.634 & 0.721 & 0.594 & 0.728 & 0.851 & 0.878 & 0.714 & 0.774 & 0.766 & 0.828 & 0.788 & 0.833 \\
& & {\scriptsize Gemini-2.0}  & 0.650 & 0.704 & 0.703 & 0.664 & 0.876 & 0.864 & 0.723 & 0.742 & 0.734 & 0.781 & 0.848 & 0.814 \\
& & {\scriptsize Qwen}       & 0.651 & 0.704 & 0.687 & 0.739 & 0.847 & 0.901 & 0.742 & 0.842 & 0.757 & 0.801 & 0.835 & 0.860 \\
\cmidrule(lr){3-15}
& \multirow{4}{*}{\rotatebox{90}{\textbf{\scriptsize Enhanced}}}
& {\scriptsize GPT-4o-mini} & 0.629 & 0.690 & 0.767 & 0.783 & 0.855 & 0.864 & 0.763 & 0.863 & 0.652 & 0.761 & 0.813 & 0.887 \\
& & {\scriptsize GPT-3.5-turbo}  & 0.620 & 0.694 & 0.684 & 0.760 & 0.841 & 0.862 & 0.700 & 0.807 & 0.564 & 0.793 & 0.667 & 0.850 \\
& & {\scriptsize Gemini-2.0}  & 0.630 & 0.700 & 0.739 & 0.752 & 0.852 & 0.879 & 0.727 & 0.830 & 0.668 & 0.789 & 0.809 & 0.865 \\
& & {\scriptsize Qwen}        & 0.628 & 0.737 & 0.747 & 0.811 & 0.847 & 0.838 & 0.698 & 0.867 & 0.650 & 0.843 & 0.759 & 0.841 \\
\bottomrule
\end{tabular}

\vspace{5pt}

\begin{tabular}{|cclcccccccccccc|}
\toprule
& & \textbf{\textbf{\scriptsize Model}} &
\multicolumn{2}{c}{{\scriptsize Dim 1 (+)}} &
\multicolumn{2}{c}{{\scriptsize Dim 1 (-)}} &
\multicolumn{2}{c}{{\scriptsize Dim 2 (+)}} &
\multicolumn{2}{c}{{\scriptsize Dim 2 (-)}} &
\multicolumn{2}{c}{{\scriptsize Dim 3 (+)}} &
\multicolumn{2}{c|}{{\scriptsize Dim 3 (-)}} \\[3pt]
\cmidrule(lr){4-5}
\cmidrule(lr){6-7}
\cmidrule(lr){8-9}
\cmidrule(lr){10-11}
\cmidrule(lr){12-13}
\cmidrule(lr){14-15}
& & & {\scriptsize LLM} & {\scriptsize RAG} & {\scriptsize LLM} & {\scriptsize RAG} & {\scriptsize LLM} & {\scriptsize RAG} & {\scriptsize LLM} & {\scriptsize RAG} &{\scriptsize LLM} & {\scriptsize RAG} & {\scriptsize LLM} & {\scriptsize RAG} \\
\midrule
\multirow{8}{*}{\rotatebox{90}{\textbf{\small Temperature 0.3}}}
& \multirow{4}{*}{\rotatebox{90}{\textbf{\scriptsize Regular}}}
& {\scriptsize GPT-4o-mini} & 0.704 & 0.710 & 0.716 & 0.724 & 0.887 & 0.892 & 0.836 & 0.830 & 0.803 & 0.799 & 0.854 & 0.847 \\
& & {\scriptsize GPT-3.5-turbo}  & 0.714 & 0.716 & 0.729 & 0.713 & 0.880 & 0.883 & 0.783 & 0.765 & 0.832 & 0.832 & 0.817 & 0.827 \\
& & {\scriptsize Gemini-2.0}   & 0.663 & 0.720 & 0.704 & 0.644 & 0.879 & 0.853 & 0.719 & 0.756 & 0.731 & 0.782 & 0.854 & 0.809 \\
& & {\scriptsize Qwen}         & 0.741 & 0.743 & 0.820 & 0.806 & 0.844 & 0.837 & 0.887 & 0.861 & 0.853 & 0.859 & 0.855 & 0.854 \\
\cmidrule(lr){3-15}
& \multirow{4}{*}{\rotatebox{90}{\textbf{\scriptsize Enhanced}}}
& {\scriptsize GPT-4o-mini}  & 0.629 & 0.680 & 0.776 & 0.794 & 0.854 & 0.870 & 0.764 & 0.864 & 0.651 & 0.767 & 0.817 & 0.884 \\
& & {\scriptsize GPT-3.5-turbo}  & 0.635 & 0.692 & 0.710 & 0.778 & 0.845 & 0.875 & 0.699 & 0.808 & 0.566 & 0.795 & 0.653 & 0.855 \\
& & {\scriptsize Gemini-2.0}  & 0.644 & 0.704 & 0.757 & 0.766 & 0.854 & 0.896 & 0.733 & 0.864 & 0.680 & 0.801 & 0.818 & 0.875 \\
& & {\scriptsize Qwen}         & 0.639 & 0.743 & 0.751 & 0.808 & 0.857 & 0.849 & 0.696 & 0.885 & 0.646 & 0.857 & 0.778 & 0.853 \\
\bottomrule
\end{tabular}

\vspace{5pt}

\begin{tabular}{|cclcccccccccccc|}
\toprule
& & \textbf{\textbf{\scriptsize Model}} &
\multicolumn{2}{c}{{\scriptsize Dim 1 (+)}} &
\multicolumn{2}{c}{{\scriptsize Dim 1 (-)}} &
\multicolumn{2}{c}{{\scriptsize Dim 2 (+)}} &
\multicolumn{2}{c}{{\scriptsize Dim 2 (-)}} &
\multicolumn{2}{c}{{\scriptsize Dim 3 (+)}} &
\multicolumn{2}{c|}{{\scriptsize Dim 3 (-)}} \\[3pt]
\cmidrule(lr){4-5}
\cmidrule(lr){6-7}
\cmidrule(lr){8-9}
\cmidrule(lr){10-11}
\cmidrule(lr){12-13}
\cmidrule(lr){14-15}
& & & {\scriptsize LLM} & {\scriptsize RAG} & {\scriptsize LLM} & {\scriptsize RAG} & {\scriptsize LLM} & {\scriptsize RAG} & {\scriptsize LLM} & {\scriptsize RAG} &{\scriptsize LLM} & {\scriptsize RAG} & {\scriptsize LLM} & {\scriptsize RAG} \\
\midrule
\multirow{8}{*}{\rotatebox{90}{\textbf{\small Temperature 0.5}}}
& \multirow{4}{*}{\rotatebox{90}{\textbf{\scriptsize Regular}}}
& {\scriptsize GPT-4o-mini} & 0.672 & 0.706 & 0.661 & 0.729 & 0.882 & 0.893 & 0.792 & 0.830 & 0.767 & 0.803 & 0.860 & 0.853 \\
& & {\scriptsize GPT-3.5-turbo}  & 0.641 & 0.711 & 0.633 & 0.733 & 0.865 & 0.880 & 0.712 & 0.776 & 0.766 & 0.835 & 0.799 & 0.841 \\
& & {\scriptsize Gemini-2.0}   & 0.662 & 0.698 & 0.679 & 0.680 & 0.868 & 0.876 & 0.723 & 0.760 & 0.742 & 0.792 & 0.844 & 0.797 \\
& & {\scriptsize Qwen}           & 0.667 & 0.739 & 0.677 & 0.815 & 0.870 & 0.836 & 0.736 & 0.884 & 0.765 & 0.868 & 0.827 & 0.855 \\
\cmidrule(lr){3-15}
& \multirow{4}{*}{\rotatebox{90}{\textbf{\scriptsize Enhanced}}}
& {\scriptsize GPT-4o-mini} & 0.645 & 0.692 & 0.775 & 0.796 & 0.854 & 0.876 & 0.767 & 0.867 & 0.659 & 0.769 & 0.815 & 0.885 \\
& & {\scriptsize GPT-3.5-turbo}  & 0.625 & 0.709 & 0.706 & 0.775 & 0.845 & 0.870 & 0.709 & 0.832 & 0.591 & 0.806 & 0.658 & 0.865 \\
& & {\scriptsize Gemini-2.0}  & 0.629 & 0.724 & 0.747 & 0.767 & 0.857 & 0.891 & 0.719 & 0.857 & 0.683 & 0.804 & 0.793 & 0.890 \\
& & {\scriptsize Qwen}        & 0.632 & 0.740 & 0.739 & 0.824 & 0.861 & 0.839 & 0.717 & 0.880 & 0.622 & 0.862 & 0.771 & 0.854 \\
\bottomrule
\end{tabular}

\vspace{5pt}

\begin{tabular}{|cclcccccccccccc|}
\toprule
& & \textbf{\textbf{\scriptsize Model}} &
\multicolumn{2}{c}{{\scriptsize Dim 1 (+)}} &
\multicolumn{2}{c}{{\scriptsize Dim 1 (-)}} &
\multicolumn{2}{c}{{\scriptsize Dim 2 (+)}} &
\multicolumn{2}{c}{{\scriptsize Dim 2 (-)}} &
\multicolumn{2}{c}{{\scriptsize Dim 3 (+)}} &
\multicolumn{2}{c|}{{\scriptsize Dim 3 (-)}} \\[3pt]
\cmidrule(lr){4-5}
\cmidrule(lr){6-7}
\cmidrule(lr){8-9}
\cmidrule(lr){10-11}
\cmidrule(lr){12-13}
\cmidrule(lr){14-15}
& & & {\scriptsize LLM} & {\scriptsize RAG} & {\scriptsize LLM} & {\scriptsize RAG} & {\scriptsize LLM} & {\scriptsize RAG} & {\scriptsize LLM} & {\scriptsize RAG} &{\scriptsize LLM} & {\scriptsize RAG} & {\scriptsize LLM} & {\scriptsize RAG} \\
\midrule
\multirow{8}{*}{\rotatebox{90}{\textbf{\small Temperature 0.7}}}
& \multirow{4}{*}{\rotatebox{90}{\textbf{\scriptsize Regular}}}
& {\scriptsize GPT-4o-mini} & 0.669 & 0.714 & 0.670 & 0.722 & 0.884 & 0.891 & 0.784 & 0.840 & 0.777 & 0.806 & 0.841 & 0.851 \\
& & {\scriptsize GPT-3.5-turbo}  & 0.639 & 0.713 & 0.634 & 0.746 & 0.857 & 0.885 & 0.697 & 0.787 & 0.754 & 0.826 & 0.809 & 0.831 \\
& & {\scriptsize Gemini-2.0}   & 0.654 & 0.693 & 0.684 & 0.668 & 0.870 & 0.873 & 0.732 & 0.756 & 0.731 & 0.811 & 0.846 & 0.804 \\
& & {\scriptsize Qwen}         & 0.663 & 0.736 & 0.674 & 0.810 & 0.863 & 0.835 & 0.732 & 0.887 & 0.769 & 0.875 & 0.831 & 0.843 \\
\cmidrule(lr){3-15}
& \multirow{4}{*}{\rotatebox{90}{\textbf{\scriptsize Enhanced}}}
& {\scriptsize GPT-4o-mini}  & 0.641 & 0.693 & 0.776 & 0.794 & 0.857 & 0.878 & 0.780 & 0.873 & 0.665 & 0.782 & 0.819 & 0.879 \\
& & {\scriptsize GPT-3.5-turbo}  & 0.625 & 0.710 & 0.707 & 0.781 & 0.836 & 0.876 & 0.709 & 0.804 & 0.579 & 0.802 & 0.644 & 0.854 \\
& & {\scriptsize Gemini-2.0}  & 0.640 & 0.725 & 0.756 & 0.762 & 0.865 & 0.884 & 0.721 & 0.846 & 0.688 & 0.789 & 0.818 & 0.890 \\
& & {\scriptsize Qwen}         & 0.632 & 0.740 & 0.747 & 0.818 & 0.848 & 0.831 & 0.714 & 0.886 & 0.664 & 0.866 & 0.764 & 0.844 \\
\bottomrule
\end{tabular}

\vspace{5pt}

\begin{tabular}{|cclcccccccccccc|}
\toprule
& & \textbf{\textbf{\scriptsize Model}} &
\multicolumn{2}{c}{{\scriptsize Dim 1 (+)}} &
\multicolumn{2}{c}{{\scriptsize Dim 1 (-)}} &
\multicolumn{2}{c}{{\scriptsize Dim 2 (+)}} &
\multicolumn{2}{c}{{\scriptsize Dim 2 (-)}} &
\multicolumn{2}{c}{{\scriptsize Dim 3 (+)}} &
\multicolumn{2}{c|}{{\scriptsize Dim 3 (-)}} \\[3pt]
\cmidrule(lr){4-5}
\cmidrule(lr){6-7}
\cmidrule(lr){8-9}
\cmidrule(lr){10-11}
\cmidrule(lr){12-13}
\cmidrule(lr){14-15}
& & & {\scriptsize LLM} & {\scriptsize RAG} & {\scriptsize LLM} & {\scriptsize RAG} & {\scriptsize LLM} & {\scriptsize RAG} & {\scriptsize LLM} & {\scriptsize RAG} &{\scriptsize LLM} & {\scriptsize RAG} & {\scriptsize LLM} & {\scriptsize RAG} \\
\midrule
\multirow{8}{*}{\rotatebox{90}{\textbf{\small Temperature 0.9}}}
& \multirow{4}{*}{\rotatebox{90}{\textbf{\scriptsize Regular}}}

& {\scriptsize GPT-4o-mini} & 0.673 & 0.710 & 0.667 & 0.729 & 0.885 & 0.886 & 0.777 & 0.840 & 0.773 & 0.804 & 0.855 & 0.851 \\
& &{\scriptsize GPT-3.5-turbo}  & 0.653 & 0.720 & 0.648 & 0.733 & 0.849 & 0.881 & 0.703 & 0.797 & 0.749 & 0.833 & 0.799 & 0.824 \\
& & {\scriptsize Gemini-2.0}   & 0.647 & 0.709 & 0.673 & 0.700 & 0.864 & 0.865 & 0.734 & 0.754 & 0.735 & 0.787 & 0.841 & 0.800 \\
& & {\scriptsize Qwen}         & 0.660 & 0.738 & 0.661 & 0.816 & 0.845 & 0.845 & 0.744 & 0.880 & 0.761 & 0.879 & 0.820 & 0.857 \\
\cmidrule(lr){3-15}
& \multirow{4}{*}{\rotatebox{90}{\textbf{\scriptsize Enhanced}}}
& {\scriptsize GPT-4o-mini}  & 0.645 & 0.694 & 0.784 & 0.791 & 0.863 & 0.872 & 0.776 & 0.870 & 0.671 & 0.779 & 0.827 & 0.878 \\
& & {\scriptsize GPT-3.5-turbo}  & 0.642 & 0.703 & 0.716 & 0.785 & 0.827 & 0.872 & 0.713 & 0.825 & 0.615 & 0.813 & 0.686 & 0.860 \\
& & {\scriptsize Gemini-2.0}   & 0.629 & 0.723 & 0.753 & 0.778 & 0.859 & 0.893 & 0.710 & 0.835 & 0.646 & 0.817 & 0.815 & 0.880 \\
& &{\scriptsize Qwen}         & 0.633 & 0.739 & 0.737 & 0.805 & 0.851 & 0.842 & 0.705 & 0.875 & 0.653 & 0.873 & 0.781 & 0.849 \\
\bottomrule
\end{tabular}
\end{table*}
\begin{table*}[ht]
\centering
\tiny
\rowcolors{0}{}{}
\caption{Hybrid cosine similarities for Regular and Enhanced prompts across five temperatures}
\label{tab:hybrid_alltemps}
\renewcommand{\arraystretch}{1.15}

\begin{tabular}{|cclcccccccccccc|}
\toprule
& & \textbf{\textbf{\scriptsize Model}} &
\multicolumn{2}{c}{{\scriptsize Dim 1 (+)}} &
\multicolumn{2}{c}{{\scriptsize Dim 1 (-)}} &
\multicolumn{2}{c}{{\scriptsize Dim 2 (+)}} &
\multicolumn{2}{c}{{\scriptsize Dim 2 (-)}} &
\multicolumn{2}{c}{{\scriptsize Dim 3 (+)}} &
\multicolumn{2}{c|}{{\scriptsize Dim 3 (-)}} \\[3pt]
\cmidrule(lr){4-5}
\cmidrule(lr){6-7}
\cmidrule(lr){8-9}
\cmidrule(lr){10-11}
\cmidrule(lr){12-13}
\cmidrule(lr){14-15}
& & & {\scriptsize LLM} & {\scriptsize RAG} & {\scriptsize LLM} & {\scriptsize RAG} & {\scriptsize LLM} & {\scriptsize RAG} & {\scriptsize LLM} & {\scriptsize RAG} &{\scriptsize LLM} & {\scriptsize RAG} & {\scriptsize LLM} & {\scriptsize RAG} \\
\midrule
\multirow{8}{*}{\rotatebox{90}{\textbf{\small Temperature 0.1}}}
& \multirow{4}{*}{\rotatebox{90}{\textbf{\scriptsize Regular}}}
& {\scriptsize GPT-4o-mini}   & 0.731 & 0.776 & 0.702 & 0.737 & 0.833 & 0.836 & 0.781 & 0.823 & 0.769 & 0.818 & 0.798 & 0.803 \\
& & {\scriptsize GPT-3.5-turbo} & 0.702 & 0.788 & 0.647 & 0.770 & 0.811 & 0.819 & 0.748 & 0.783 & 0.788 & 0.857 & 0.746 & 0.781 \\
& & {\scriptsize Gemini-2.0}    & 0.727 & 0.785 & 0.715 & 0.735 & 0.832 & 0.844 & 0.740 & 0.785 & 0.759 & 0.828 & 0.796 & 0.788 \\
& & {\scriptsize Qwen}          & 0.726 & 0.783 & 0.718 & 0.775 & 0.824 & 0.854 & 0.775 & 0.856 & 0.783 & 0.829 & 0.793 & 0.815 \\
\cmidrule(lr){3-15}
& \multirow{4}{*}{\rotatebox{90}{\textbf{\scriptsize Enhanced}}}
& {\scriptsize GPT-4o-mini}   & 0.690 & 0.747 & 0.749 & 0.798 & 0.721 & 0.746 & 0.750 & 0.855 & 0.725 & 0.806 & 0.785 & 0.847 \\
& & {\scriptsize GPT-3.5-turbo} & 0.692 & 0.770 & 0.720 & 0.800 & 0.754 & 0.793 & 0.716 & 0.809 & 0.673 & 0.842 & 0.701 & 0.821 \\
& & {\scriptsize Gemini-2.0}   & 0.697 & 0.782 & 0.739 & 0.793 & 0.744 & 0.821 & 0.717 & 0.843 & 0.718 & 0.844 & 0.781 & 0.840 \\
& & {\scriptsize Qwen}          & 0.697 & 0.821 & 0.752 & 0.857 & 0.728 & 0.803 & 0.710 & 0.879 & 0.713 & 0.879 & 0.761 & 0.805 \\
\bottomrule
\end{tabular}

\vspace{5pt}


\begin{tabular}{|cclcccccccccccc|}
\toprule
& & \textbf{\textbf{\scriptsize Model}} &
\multicolumn{2}{c}{{\scriptsize Dim 1 (+)}} &
\multicolumn{2}{c}{{\scriptsize Dim 1 (-)}} &
\multicolumn{2}{c}{{\scriptsize Dim 2 (+)}} &
\multicolumn{2}{c}{{\scriptsize Dim 2 (-)}} &
\multicolumn{2}{c}{{\scriptsize Dim 3 (+)}} &
\multicolumn{2}{c|}{{\scriptsize Dim 3 (-)}} \\[3pt]
\cmidrule(lr){4-5}
\cmidrule(lr){6-7}
\cmidrule(lr){8-9}
\cmidrule(lr){10-11}
\cmidrule(lr){12-13}
\cmidrule(lr){14-15}
& & & {\scriptsize LLM} & {\scriptsize RAG} & {\scriptsize LLM} & {\scriptsize RAG} & {\scriptsize LLM} & {\scriptsize RAG} & {\scriptsize LLM} & {\scriptsize RAG} &{\scriptsize LLM} & {\scriptsize RAG} & {\scriptsize LLM} & {\scriptsize RAG} \\
\midrule
\multirow{8}{*}{\rotatebox{90}{\textbf{\small Temperature 0.3}}}
& \multirow{4}{*}{\rotatebox{90}{\textbf{\scriptsize Regular}}}
& {\scriptsize GPT-4o-mini}  & 0.776 & 0.780 & 0.750 & 0.755 & 0.836 & 0.836 & 0.830 & 0.821 & 0.821 & 0.815 & 0.813 & 0.807 \\
& & {\scriptsize GPT-3.5-turbo}& 0.785 & 0.787 & 0.774 & 0.760 & 0.816 & 0.820 & 0.798 & 0.787 & 0.858 & 0.860 & 0.778 & 0.783 \\
& & {\scriptsize Gemini-2.0}    & 0.734 & 0.794 & 0.715 & 0.713 & 0.836 & 0.817 & 0.737 & 0.777 & 0.759 & 0.828 & 0.800 & 0.783 \\
& & {\scriptsize Qwen}          & 0.821 & 0.822 & 0.859 & 0.852 & 0.805 & 0.807 & 0.898 & 0.877 & 0.886 & 0.890 & 0.818 & 0.815 \\
\cmidrule(lr){3-15}
& \multirow{4}{*}{\rotatebox{90}{\textbf{\scriptsize Enhanced}}}
& {\scriptsize GPT-4o-mini}   & 0.689 & 0.739 & 0.763 & 0.806 & 0.727 & 0.753 & 0.752 & 0.853 & 0.720 & 0.809 & 0.785 & 0.844 \\
& & {\scriptsize GPT-3.5-turbo} & 0.709 & 0.767 & 0.734 & 0.809 & 0.755 & 0.807 & 0.721 & 0.808 & 0.685 & 0.841 & 0.688 & 0.831 \\
& & {\scriptsize Gemini-2.0}   & 0.698 & 0.790 & 0.740 & 0.801 & 0.741 & 0.839 & 0.725 & 0.864 & 0.727 & 0.847 & 0.796 & 0.851 \\
& & {\scriptsize Qwen}    & 0.705 & 0.828 & 0.751 & 0.853 & 0.743 & 0.809 & 0.718 & 0.892 & 0.712 & 0.887 & 0.776 & 0.810 \\
\bottomrule
\end{tabular}

\vspace{5pt}

\begin{tabular}{|cclcccccccccccc|}
\toprule
& & \textbf{\textbf{\scriptsize Model}} &
\multicolumn{2}{c}{{\scriptsize Dim 1 (+)}} &
\multicolumn{2}{c}{{\scriptsize Dim 1 (-)}} &
\multicolumn{2}{c}{{\scriptsize Dim 2 (+)}} &
\multicolumn{2}{c}{{\scriptsize Dim 2 (-)}} &
\multicolumn{2}{c}{{\scriptsize Dim 3 (+)}} &
\multicolumn{2}{c|}{{\scriptsize Dim 3 (-)}} \\[3pt]
\cmidrule(lr){4-5}
\cmidrule(lr){6-7}
\cmidrule(lr){8-9}
\cmidrule(lr){10-11}
\cmidrule(lr){12-13}
\cmidrule(lr){14-15}
& & & {\scriptsize LLM} & {\scriptsize RAG} & {\scriptsize LLM} & {\scriptsize RAG} & {\scriptsize LLM} & {\scriptsize RAG} & {\scriptsize LLM} & {\scriptsize RAG} &{\scriptsize LLM} & {\scriptsize RAG} & {\scriptsize LLM} & {\scriptsize RAG} \\
\midrule
\multirow{8}{*}{\rotatebox{90}{\textbf{\small Temperature 0.5}}}
& \multirow{4}{*}{\rotatebox{90}{\textbf{\scriptsize Regular}}}
&  {\scriptsize GPT-4o-mini}   & 0.734 & 0.772 & 0.693 & 0.752 & 0.827 & 0.838 & 0.787 & 0.819 & 0.778 & 0.817 & 0.804 & 0.806 \\
& & {\scriptsize GPT-3.5-turbo} & 0.704 & 0.781 & 0.673 & 0.774 & 0.815 & 0.820 & 0.750 & 0.804 & 0.783 & 0.865 & 0.761 & 0.787 \\
& & {\scriptsize Gemini-2.0}   & 0.735 & 0.781 & 0.699 & 0.738 & 0.826 & 0.832 & 0.737 & 0.790 & 0.762 & 0.832 & 0.795 & 0.782 \\
& & {\scriptsize Qwen}    & 0.741 & 0.823 & 0.710 & 0.858 & 0.837 & 0.804 & 0.773 & 0.888 & 0.786 & 0.896 & 0.794 & 0.815 \\
\cmidrule(lr){3-15}
& \multirow{4}{*}{\rotatebox{90}{\textbf{\scriptsize Enhanced}}}
& {\scriptsize GPT-4o-mini}  & 0.695 & 0.744 & 0.756 & 0.807 & 0.731 & 0.758 & 0.754 & 0.857 & 0.723 & 0.809 & 0.785 & 0.847 \\
& & {\scriptsize GPT-3.5-turbo}  & 0.702 & 0.785 & 0.728 & 0.810 & 0.742 & 0.808 & 0.728 & 0.829 & 0.694 & 0.849 & 0.691 & 0.833 \\
& & {\scriptsize Gemini-2.0}    & 0.691 & 0.802 & 0.738 & 0.804 & 0.743 & 0.844 & 0.721 & 0.866 & 0.723 & 0.857 & 0.776 & 0.853 \\
& & {\scriptsize Qwen}  & 0.703 & 0.823 & 0.743 & 0.862 & 0.736 & 0.806 & 0.730 & 0.891 & 0.693 & 0.891 & 0.764 & 0.818 \\
\bottomrule
\end{tabular}

\vspace{5pt}

\begin{tabular}{|cclcccccccccccc|}
\toprule
& & \textbf{\textbf{\scriptsize Model}} &
\multicolumn{2}{c}{{\scriptsize Dim 1 (+)}} &
\multicolumn{2}{c}{{\scriptsize Dim 1 (-)}} &
\multicolumn{2}{c}{{\scriptsize Dim 2 (+)}} &
\multicolumn{2}{c}{{\scriptsize Dim 2 (-)}} &
\multicolumn{2}{c}{{\scriptsize Dim 3 (+)}} &
\multicolumn{2}{c|}{{\scriptsize Dim 3 (-)}} \\[3pt]
\cmidrule(lr){4-5}
\cmidrule(lr){6-7}
\cmidrule(lr){8-9}
\cmidrule(lr){10-11}
\cmidrule(lr){12-13}
\cmidrule(lr){14-15}
& & & {\scriptsize LLM} & {\scriptsize RAG} & {\scriptsize LLM} & {\scriptsize RAG} & {\scriptsize LLM} & {\scriptsize RAG} & {\scriptsize LLM} & {\scriptsize RAG} &{\scriptsize LLM} & {\scriptsize RAG} & {\scriptsize LLM} & {\scriptsize RAG} \\
\midrule
\multirow{8}{*}{\rotatebox{90}{\textbf{\small Temperature 0.7}}}
& \multirow{4}{*}{\rotatebox{90}{\textbf{\scriptsize Regular}}}
&  {\scriptsize GPT-4o-mini}  & 0.740 & 0.780 & 0.690 & 0.750 & 0.830 & 0.840 & 0.790 & 0.830 & 0.790 & 0.820 & 0.790 & 0.800 \\
& & {\scriptsize GPT-3.5-turbo} & 0.710 & 0.780 & 0.680 & 0.790 & 0.820 & 0.830 & 0.740 & 0.810 & 0.780 & 0.850 & 0.770 & 0.780 \\
& & {\scriptsize Gemini-2.0}    & 0.730 & 0.780 & 0.700 & 0.730 & 0.830 & 0.830 & 0.740 & 0.790 & 0.750 & 0.850 & 0.800 & 0.780 \\
& & {\scriptsize Qwen}          & 0.740 & 0.820 & 0.710 & 0.850 & 0.830 & 0.800 & 0.780 & 0.900 & 0.790 & 0.900 & 0.800 & 0.810 \\
\cmidrule(lr){3-15}
& \multirow{4}{*}{\rotatebox{90}{\textbf{\scriptsize Enhanced}}}
&  {\scriptsize GPT-4o-mini}   & 0.690 & 0.750 & 0.760 & 0.800 & 0.730 & 0.770 & 0.760 & 0.860 & 0.720 & 0.820 & 0.790 & 0.840 \\
& & {\scriptsize GPT-3.5-turbo} & 0.690 & 0.780 & 0.720 & 0.810 & 0.730 & 0.800 & 0.720 & 0.820 & 0.690 & 0.850 & 0.690 & 0.820 \\
& &  {\scriptsize Gemini-2.0}   & 0.700 & 0.800 & 0.740 & 0.800 & 0.750 & 0.840 & 0.720 & 0.850 & 0.730 & 0.850 & 0.790 & 0.860 \\
& & {\scriptsize Qwen}       & 0.700 & 0.820 & 0.750 & 0.860 & 0.740 & 0.800 & 0.730 & 0.890 & 0.720 & 0.900 & 0.760 & 0.810 \\
\bottomrule
\end{tabular}

\vspace{5pt}


\begin{tabular}{|cclcccccccccccc|}
\toprule
& & \textbf{\textbf{\scriptsize Model}} &
\multicolumn{2}{c}{{\scriptsize Dim 1 (+)}} &
\multicolumn{2}{c}{{\scriptsize Dim 1 (-)}} &
\multicolumn{2}{c}{{\scriptsize Dim 2 (+)}} &
\multicolumn{2}{c}{{\scriptsize Dim 2 (-)}} &
\multicolumn{2}{c}{{\scriptsize Dim 3 (+)}} &
\multicolumn{2}{c|}{{\scriptsize Dim 3 (-)}} \\[3pt]
\cmidrule(lr){4-5}
\cmidrule(lr){6-7}
\cmidrule(lr){8-9}
\cmidrule(lr){10-11}
\cmidrule(lr){12-13}
\cmidrule(lr){14-15}
& & & {\scriptsize LLM} & {\scriptsize RAG} & {\scriptsize LLM} & {\scriptsize RAG} & {\scriptsize LLM} & {\scriptsize RAG} & {\scriptsize LLM} & {\scriptsize RAG} &{\scriptsize LLM} & {\scriptsize RAG} & {\scriptsize LLM} & {\scriptsize RAG} \\
\midrule
\multirow{8}{*}{\rotatebox{90}{\textbf{\small Temperature 0.9}}}
& \multirow{4}{*}{\rotatebox{90}{\textbf{\scriptsize Regular}}}

&  {\scriptsize GPT-4o-mini}  & 0.737 & 0.778 & 0.697 & 0.759 & 0.831 & 0.830 & 0.782 & 0.836 & 0.785 & 0.821 & 0.805 & 0.811 \\
& & {\scriptsize GPT-3.5-turbo}& 0.716 & 0.786 & 0.681 & 0.779 & 0.813 & 0.829 & 0.742 & 0.812 & 0.772 & 0.860 & 0.762 & 0.778 \\
& &  {\scriptsize Gemini-2.0}     & 0.727 & 0.785 & 0.693 & 0.754 & 0.823 & 0.836 & 0.744 & 0.773 & 0.759 & 0.828 & 0.790 & 0.776 \\
& & {\scriptsize Qwen}         & 0.738 & 0.822 & 0.709 & 0.858 & 0.816 & 0.807 & 0.776 & 0.887 & 0.787 & 0.906 & 0.790 & 0.816 \\
\cmidrule(lr){3-15}
& \multirow{4}{*}{\rotatebox{90}{\textbf{\scriptsize Enhanced}}}
&  {\scriptsize GPT-4o-mini}  & 0.698 & 0.750 & 0.758 & 0.801 & 0.732 & 0.763 & 0.761 & 0.855 & 0.732 & 0.816 & 0.790 & 0.845 \\
& & {\scriptsize GPT-3.5-turbo}& 0.714 & 0.771 & 0.726 & 0.819 & 0.728 & 0.808 & 0.731 & 0.820 & 0.706 & 0.856 & 0.711 & 0.830 \\
& &  {\scriptsize Gemini-2.0}   & 0.692 & 0.800 & 0.737 & 0.807 & 0.735 & 0.851 & 0.705 & 0.855 & 0.696 & 0.861 & 0.793 & 0.852 \\
& &{\scriptsize Qwen}     & 0.706 & 0.823 & 0.742 & 0.853 & 0.733 & 0.806 & 0.721 & 0.886 & 0.712 & 0.900 & 0.773 & 0.813 \\
\bottomrule
\end{tabular}
\end{table*}

\end{document}